\documentclass{article}

\newcommand{\myparagraph}[1]{
\noindent
\textbf{#1} ---
}
\PassOptionsToPackage{numbers, compress}{natbib}
\usepackage{pifont}
\usepackage{wrapfig}
\usepackage{mwe}
\usepackage[dvipsnames]{xcolor}
\definecolor{BrickRed}{RGB}{178,34,34}
\definecolor{RoyalBlue}{RGB}{24,116,205}
\definecolor{OliveGreen}{RGB}{59,109,17}
\usepackage{amsmath}
 \usepackage[main, final]{neurips_2026}
\usepackage[utf8]{inputenc} % allow utf-8 input
\usepackage[T1]{fontenc}    % use 8-bit T1 fonts
\usepackage{hyperref}       % hyperlinks
\usepackage{url}            % simple URL typesetting
\usepackage{booktabs}       % professional-quality tables
\usepackage{amsfonts}       % blackboard math symbols
\usepackage{nicefrac}       % compact symbols for 1/2, etc.
\usepackage{microtype}      % microtypography
\usepackage{xcolor}         % colors
\usepackage{dsfont}
\usepackage{subcaption}
\usepackage{adjustbox}
\title{Scaling Multi-Teacher Distillation for Digital Pathology}

\author{
  Sofiène Boutaj$^{{1,2}^{*}}$, Pierre Marza$^{{1,2}^{*}}$, Varun Belagali$^{3}$, Dimitris Samaras$^{3}$,\\
  \textbf{Maria Vakalopoulou}$^{1,2}$\textbf{, Stergios Christodoulidis}$^{1,2}$\\
  \\
  $^1$ Université Paris-Saclay, CentraleSupélec, Gustave Roussy, INSERM, \\ IHU PRISM, Cancer Data Science Unit, France \\
  $^2$ Université Paris-Saclay, CentraleSupélec, MICS Laboratory, France \\
  $^3$ Department of Computer Science, Stony Brook University, USA \\
}

\begin{document}

\maketitle
\def\thefootnote{*}\footnotetext{denotes equal contribution.}
\def\thefootnote{}\footnotetext{Corresponding authors: \href{mailto:sofiene.boutaj@centralesupelec.fr}{sofiene.boutaj@centralesupelec.fr}, \href{mailto:pierre.marza@centralesupelec.fr}{pierre.marza@centralesupelec.fr}}

\begin{center}
\vspace{-0.4cm}
    \href{https://sofiene-boutaj.github.io/omni-projectpage/}{%
        \raisebox{-0.35\height}{\includegraphics[height=1.7\baselineskip]{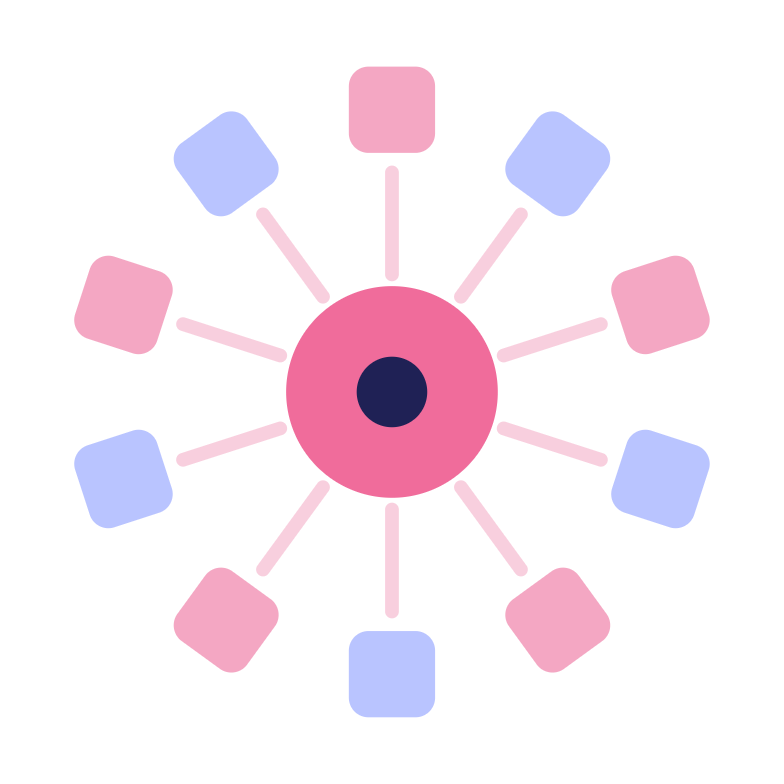}}%
        \hspace{0.3em}{\large Project page}%
    }%
    \hspace{0.2in}%
    \href{https://github.com/prism-oncology/omni}{%
        \raisebox{-0.35\height}{\includegraphics[height=1.7\baselineskip]{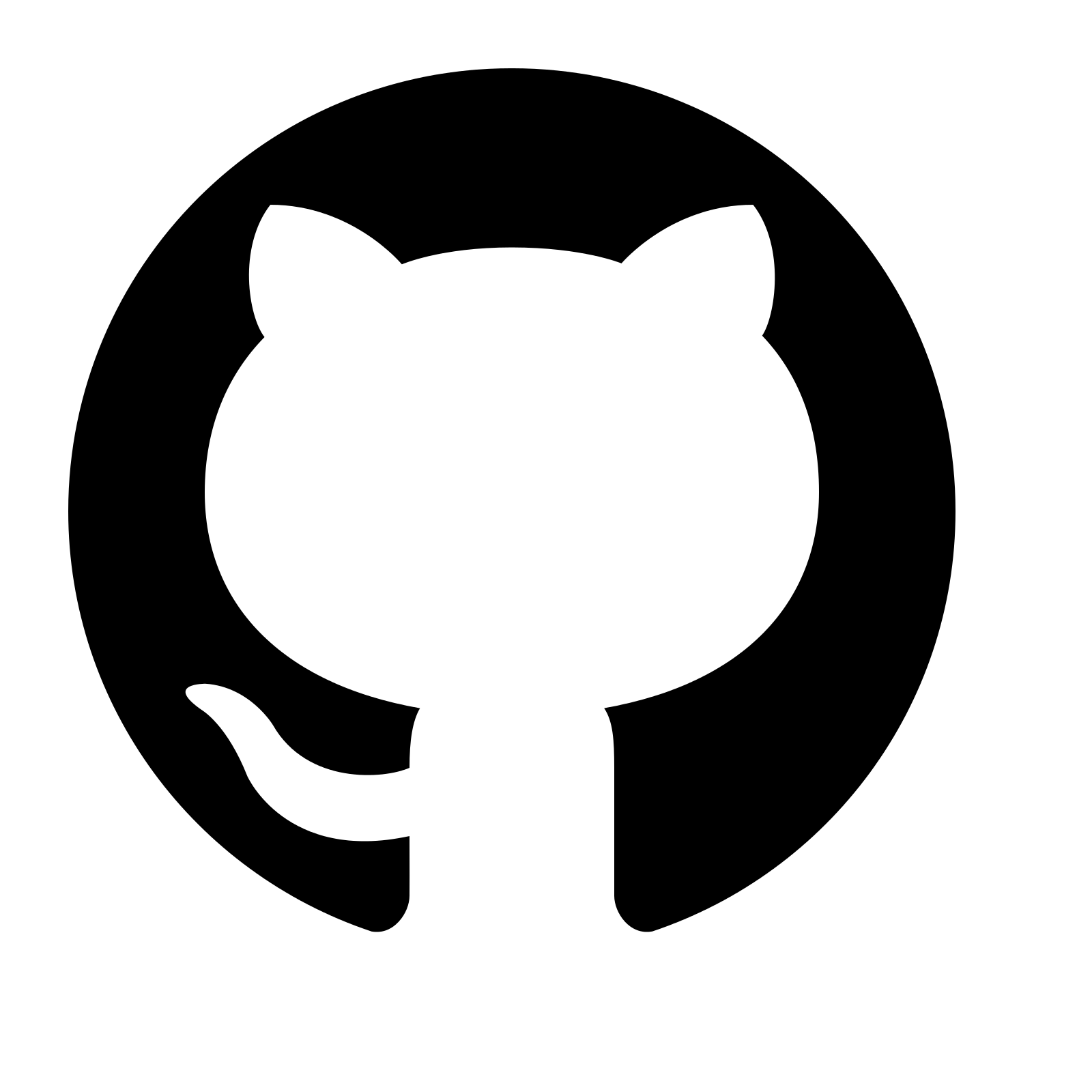}}%
        \hspace{0.3em}{\large Github}%
    }%
    \hspace{0.2in}%
    \href{https://huggingface.co/collections/sofieneb/omni-models}{%
        \raisebox{-0.35\height}{\includegraphics[height=1.7\baselineskip]{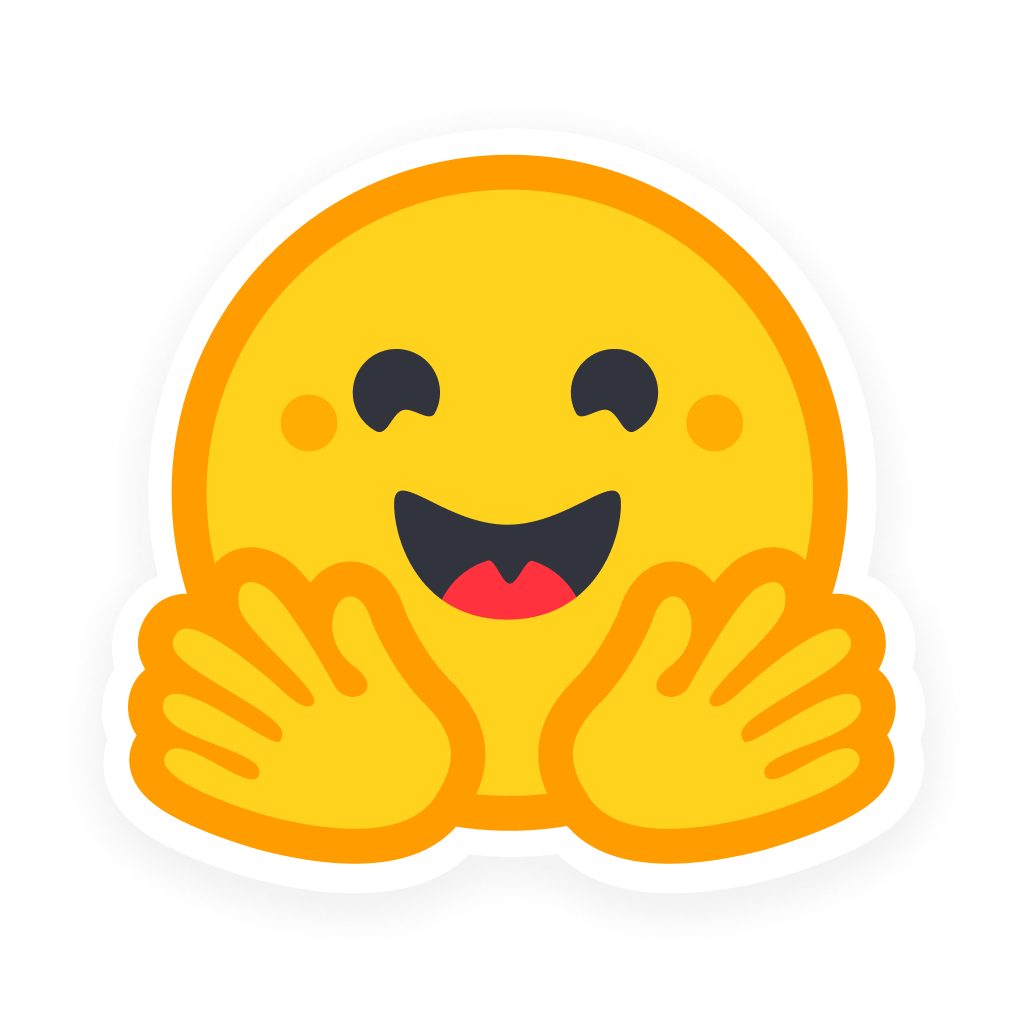}}%
        \hspace{0.3em}{\large Models}%
    }
    \hspace{0.2in}%
    \href{https://huggingface.co/collections/sofieneb/pathology-foundation-model-features}{%
        \raisebox{-0.35\height}{\includegraphics[height=1.7\baselineskip]{res/hf_logo.png}}%
        \hspace{0.3em}{\large Datasets}%
    }
    \vspace{0.4cm}
\end{center}

\begin{abstract}
  Multi-teacher distillation allows transferring knowledge from multiple ``teacher'' networks to a single ``student'' network. This method is promising in fields such as digital pathology, where many powerful foundation models were proposed recently. However, as we show in this paper, the standard multi-teacher distillation approach reacts poorly to an increase in the number of teachers used to train a student encoder. We demonstrate that learning teacher-specific representations is a key to scaling in MTD. Importantly, different design choices, such as learnable teacher tokens, a tailored attention scheme, additional mixture-of-experts layers and a contrastive loss, are proposed to better learn such teacher-specific representations. We show that our method better scales with respect to the number of considered teachers, allowing us to train compact student encoders, up to $10\times$ more efficient than larger teacher foundation models, while matching their performance and even outperforming them on a large set of tile-level and slide-level tasks. We conduct a thorough empirical validation, evaluating more than $10$ foundation models on $39$ tasks spanning tile and slide levels. As a result, we release a new collection of strong and efficient foundation models, named \textbf{OMNI}, trained from the knowledge of $10$ state-of-the-art foundation models. We finally conduct an analysis of the learned teacher-specific representations, highlighting their complementarity and explaining why they can be easily aggregated at downstream time.
\end{abstract}

\begin{figure}[t]
    \centering
    \includegraphics[width=1\linewidth]{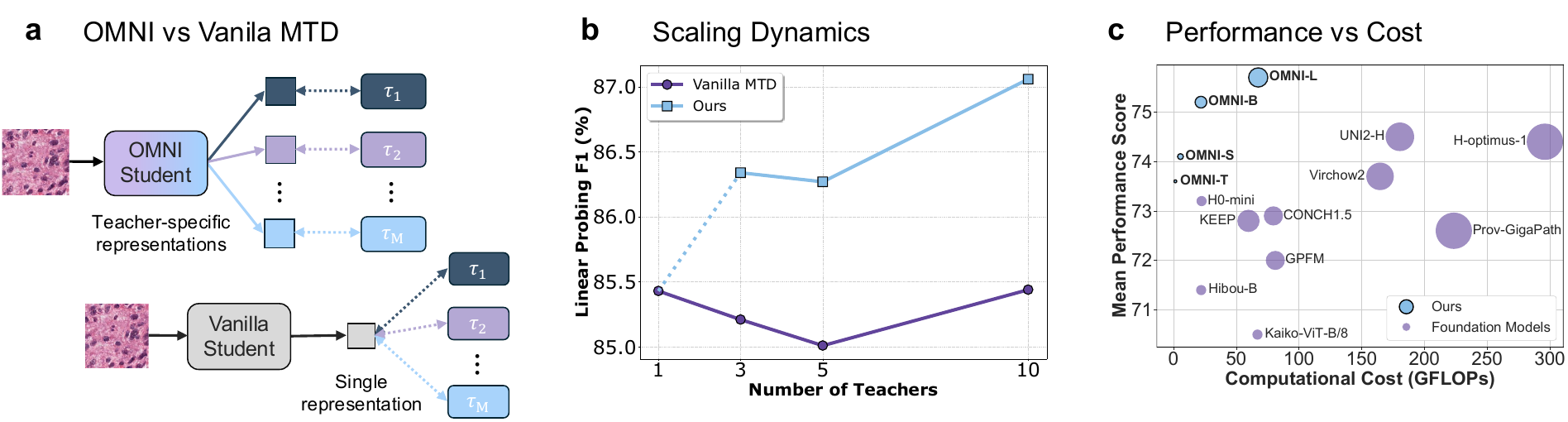}
    \caption{\textbf{OMNI}. We propose a new method to scale Multi-Teacher Distillation (MTD) to more teachers. \textbf{(a)} Unlike vanilla MTD, we learn to extract teacher-specific representations. \textbf{(b)} Downstream performance (tile-level linear probing averaged across 13 tasks) better scales when increasing the diversity of teachers for our method. For a fair evaluation, we rank teachers by their individual performance on the THUNDER benchmark and add them from strongest to weakest. \textbf{(c)} Overall we provide a collection of light vision encoder ($7-330M$ parameters) models that are competitive with and even outperform all state-of-the-art larger teacher foundation models. In particular, we report here the average of the tile-level and slide-level linear probing performances (see Table~\ref{tab:lp_tile_slide} for more details) as a function of GFLOPs. Circle sizes indicate the number of parameters of each model.}
    \label{fig:teaser}
\end{figure}

\section{Introduction}
\label{sec:intro}

Knowledge distillation~\cite{hinton2015distilling} has been extensively studied over the last few decades, yet the main goal has stayed the same throughout it: transferring knowledge from a large model, typically referred to as the teacher, to a smaller one, the student. Such knowledge distillation can either be related to a specific downstream task (supervised~\cite{hinton2015distilling}) or focus on matching a representation distribution (self-supervised~\cite{fangseed}). An extension of this line of research is Multi-Teacher Distillation (MTD), where the objective is to distill knowledge from a set of diverse teachers~\cite{Ranzinger_2024_CVPR, ranzinger2024phisdistributionbalancinglabelfree, shang2024theia, sariyildiz2025dune, sariyildiz2024unic} by fusing different representations into a powerful common latent space.

Importantly, while recent work~\cite{ranzinger2024phisdistributionbalancinglabelfree, shang2024theia, sariyildiz2025dune, sariyildiz2024unic} has improved upon the original methods~\cite{Ranzinger_2024_CVPR}, a central question has not yet been addressed in the MTD literature: \textit{What are the scaling dynamics of MTD with respect to the number of considered teachers?} Indeed, most previous work only leverages a few teachers, often $3$ or rarely $5$ of them, and does not explore the benefits of adding more teachers. 

Such a gap in the literature is mainly visible now that the diversity of existing foundation models has dramatically increased in the past few years. A typical example of this is digital pathology where, similarly to other medical imaging sub-fields, many pre-trained vision encoders were released recently~\cite{chen2024towards, vorontsov2024foundation, zimmermann2024virchow2, hoptimus0, filiot2023scaling, filiot2024phikon, nechaev2024hibou, karasikov2025training, vaidya2025molecular, wang2024pathology, xu2024whole, lu2024visual, ding2024multimodal, zhou2024knowledge, ikezogwo2023quilt, huang2023visual, xiang2025vision, shaikovski2024prism}. MTD is particularly appealing in this context and, despite some recent works exploring its application to digital pathology~\cite{ma2024towards, grashei2025pathryoshka, kusupati2022matryoshka, belagali2025ticon}, its scaling remains largely underexplored.

In this study, we thus investigate the scaling laws of MTD, with a specific focus on digital pathology foundation models. We discover, as shown in Figure~\ref{fig:teaser}b, that increasing the number of teachers for vanilla MTD (baseline) does not improve downstream performance, and even degrades it. We hypothesize that this drop in performance stems from the difficulty of reconstructing a large set of distinct embedding spaces from a single, usually small, common latent space.

Aiming in solving the scaling problem in MTD, we propose a simple yet powerful method based on teacher-specific representations (Figure~\ref{fig:teaser}a). Such a design leads to a better performance scaling with respect to the integration of more foundation models as teachers (Figure~\ref{fig:teaser}b). We show that our method is efficient at distilling up to $10$ pathology foundation models, some of which are up to $1.1B$ parameters, into smaller students ($7M - 330M$ parameters) trained on only $10M$ public TCGA tiles. Moreover, we release a new family of models, named \textbf{OMNI}, that outperform state-of-the-art larger foundation models used as teachers while being parameter-efficient (Figure~\ref{fig:teaser}c).

Our contributions can be summarized as follows:
\begin{itemize}
    \item \textbf{Learning teacher-specific representations allows scaling in MTD}: We show that vanilla MTD does not directly benefit from broadening the pool of teachers, and overcome this limitation by training our student model to extract teacher-specific representations instead of a single global representation.
    \item \textbf{A novel methodological framework supporting teacher-specific representation learning}:  We propose important design choices, such as learnable teacher embeddings, Mixture-of-Expert (MoE) layers, and a specific attention masking to better learn teacher-specific representations.
    \item \textbf{A new collection of efficient and powerful pathology foundation models}: We introduce \textbf{OMNI}, a collection of lightweight pathology vision encoders with different sizes (OMNI-T, OMNI-S, OMNI-B, OMNI-L),  distilled from $10$ of the current best state-of-the-art open-sourced foundation models. While being up to $10\times$ more efficient (GFLOPs) compared with concurrent models, they are competitive and even outperform all of them on a diverse set of $39$ tile-level and slide-level tasks. 
\end{itemize}

\section{Related work}
\label{sec:related_work}

\myparagraph{Improving the flexibility of the Transformer} The Transformer model~\cite{vaswani2017attention}, or its vision counterpart ViT~\cite{dosovitskiy2020image}, is able to adapt representations through its core attention mechanism. However, more recent work has proposed methods to improve such a flexibility on several complementary axes. A first direction consists in designing specialized attention schemes. A prominent example arises in multi-modal applications, where specific cross-attention can be applied to enrich one modality with information coming from another one~\cite{tan2019lxmert, alayrac2022flamingo, li2023blip, kim2024openvla}. Optimizing a set of input token embeddings for a specific task or domain is another way to adapt Transformer-based representations. This has been extensively explored when adapting a pre-trained Transformer to a new task: prompt tuning~\cite{liu2022p, jia2022visual, zhu2023prompt} introduces a small set of learnable tokens prepended to the input, providing a lightweight yet effective way to steer the model toward task-specific behavior. Finally, another orthogonal approach is the integration of Mixture-of-Experts (MoE) layers into Transformer-based models~\cite{shazeer2017outrageously, jiang2024mixtral, fedus2022switch}. Their main goal is to allow the activation of different experts, i.e. feed-forward networks, to modulate neural activations based on the input.

We follow this direction of making a student ViT-based model more flexible to learn diverse teacher-specific representations, revisiting the above ideas in the context of multi-teacher distillation.

\myparagraph{Multi-teacher distillation (MTD)} RADIO~\cite{Ranzinger_2024_CVPR} was one of the first papers proposing to fuse the knowledge of various pre-trained vision encoders by training a student model to decode their representations simultaneously. Theia~\cite{shang2024theia} applied this concept to robotics, showing the advantage of aggregating diverse teachers, and focused specifically on improving dense spatial representations. Following this, different works addressed arising concerns when considering diverse teachers. First, PHI-S~\cite{ranzinger2024phisdistributionbalancinglabelfree} proposed to study how to normalize representations from different teachers using diverse standardization techniques. Then, UNIC~\cite{sariyildiz2024unic} studied the impact of regularization techniques such as teacher dropping, and DUNE~\cite{sariyildiz2025dune} went one step further by considering very different teachers (2D, 3D, human pose estimation) and showed how improving the capacity of decoder heads was beneficial in such a regime. However, the current MTD frameworks include three to five teachers without studying scaling laws to more teachers.

Unlike previous work, we highlight a lack of clear scaling trends in multi-teacher distillation and propose a method to better benefit from integrating more teachers. This allows to distill knowledge from $10$ state-of-the-art foundation model teachers into a compact student vision encoder.

\myparagraph{MTD for histopathology} Multi-teacher distillation has recently been studied in the field of digital pathology to leverage existing foundation models as teachers. GPFM~\cite{ma2024towards} proposed integrating MTD with DINOv2-style~\cite{oquab2023dinov2} self-distillation pretraining. Pathryoshka~\cite{grashei2025pathryoshka} explored MTD with Matryoshka-based~\cite{kusupati2022matryoshka} nested embeddings, providing adaptable embedding dimensions in the student model. Further, TICON~\cite{belagali2025ticon} explored MTD by pretraining a slide-encoder on multiple tile-encoder embeddings with an omni-feature masked modeling approach. Current MTD frameworks in this domain have explored pretraining using only up to three teachers and have not studied the scaling trends with respect to the number of considered teachers.

\myparagraph{Pathology foundation models} are trained with self-supervised objectives to capture patterns in large-scale unsupervised histopathology datasets. We can differentiate vision-only~\cite{chen2024towards, vorontsov2024foundation, zimmermann2024virchow2, hoptimus0, filiot2023scaling, filiot2024phikon, nechaev2024hibou, karasikov2025training, vaidya2025molecular, wang2024pathology, xu2024whole, belagali2025ticon} and vision-language~\cite{lu2024visual, ding2024multimodal, zhou2024knowledge, ikezogwo2023quilt, huang2023visual, xiang2025vision, shaikovski2024prism} models. The former are trained with DiNO-style losses~\cite{caron2021emerging, oquab2023dinov2} and the latter following CLIP-like objectives~\cite{radford2021learning, yu2022coca}. As shown in previous work benchmarking pathology foundation models~\cite{wolflein2023benchmarking, kang2023benchmarking, neidlinger2024benchmarking, gustafsson2024evaluating, alfasly2024foundation, gatopoulos2024eva, jaume2024hest, breen2025comprehensive, lee2025benchmarking, majzoub2025good, alfasly2025validation, zhang2025accelerating, campanella2025clinical, ma2025pathbench, marza2025thunder}, these models have different strengths and weaknesses, and their ranking depends on the data domains and tasks to solve. Such a result indicates that they extract diverse and, thus, complementary information, mainly due to their pretraining data.

In this work, we propose to train an efficient vision model from the knowledge of a large set of open-sourced foundation models in order to leverage their diverse and complementary information.

\section{Training a vision encoder from multiple teachers}
\label{sec:method}

\subsection{Preliminaries}

\myparagraph{Vision Transformer} Our method specifically addresses the shortcomings of Vision Transformer (ViT)~\cite{dosovitskiy2020image} models trained with multi-teacher distillation. A ViT processes an image
$x \in \mathbb{R}^{H \times W \times C}$ by first splitting it into $N_p$ non-overlapping patches of size $P \times P$, yielding $N_p = HW/P^2$ patch tokens. Each patch is linearly projected to a $D$-dimensional embedding, and a single learnable \emph{CLS token} $\mathbf{c} \in \mathbb{R}^D$ is prepended, giving the input sequence,
\begin{equation}
  \mathbf{X}_0 = \left[\mathbf{c},\; \mathbf{z}^{\mathrm{patch}}_1,\;\ldots,\;
  \mathbf{z}^{\mathrm{patch}}_{N_p}\right] + \mathbf{E}^{\mathrm{pos}}
  \;\in\; \mathbb{R}^{(1+N_p)\times D},
\end{equation}
where $\mathbf{E}^{\mathrm{pos}} \in \mathbb{R}^{(1+N_p)\times D}$ are learnable positional embeddings. In the next paragraphs, we will refer to the sequence of patch tokens only as  $\mathbf{Z}^{\mathrm{patch}}$. 

$\mathbf{X}_0$ is passed through $L$ transformer blocks, each consisting of a multi-head self-attention (MHSA) and a feed-forward network (FFN),
\begin{equation}
  \mathbf{X}_\ell = \mathrm{FFN}\!\left(\mathrm{MHSA}(\mathbf{X}_{\ell-1})\right)
  + \mathbf{X}_{\ell-1}, \qquad \ell = 1,\ldots,L.
\end{equation}
In standard MHSA, every token attends to every other token, corresponding to a fully dense attention mask $\mathbf{A} = \mathbf{1}_{(1+N_p)\times(1+N_p)}$. After the final block, the CLS token $\mathbf{z}^{\mathrm{cls}} = [\mathbf{X}_L]_0 \in \mathbb{R}^D$ serves as the global image representation used for downstream tasks.

\myparagraph{Vanilla ~\cite{Ranzinger_2024_CVPR}}
Let us define a ViT-based student model $s_\theta$ parametrized by weights $\theta$ and a set of $M$ ViT teacher models $\mathcal{T} = \{t_{\psi_i}\}_{i=1...M}$, each parametrized by weights $\psi_i$. The objective is then to reconstruct, for a given input image, the representations of all teachers from that of the student. To this end, we must also introduce, for each teacher, \textit{(i)} a linear layer $p^{\mathrm{cls}}_{w_i}$ parametrized by weights $w_i$ to project the student CLS token to the teacher CLS embedding space, and \textit{(ii)} a linear layer $p^{\mathrm{patch}}_{\bar{w}_i}$ parametrized by weights $\bar{w}_i$ to project the student patch tokens to the teacher patch embedding space.

\begin{figure}[t]
  \centering
  \includegraphics[width=\linewidth]{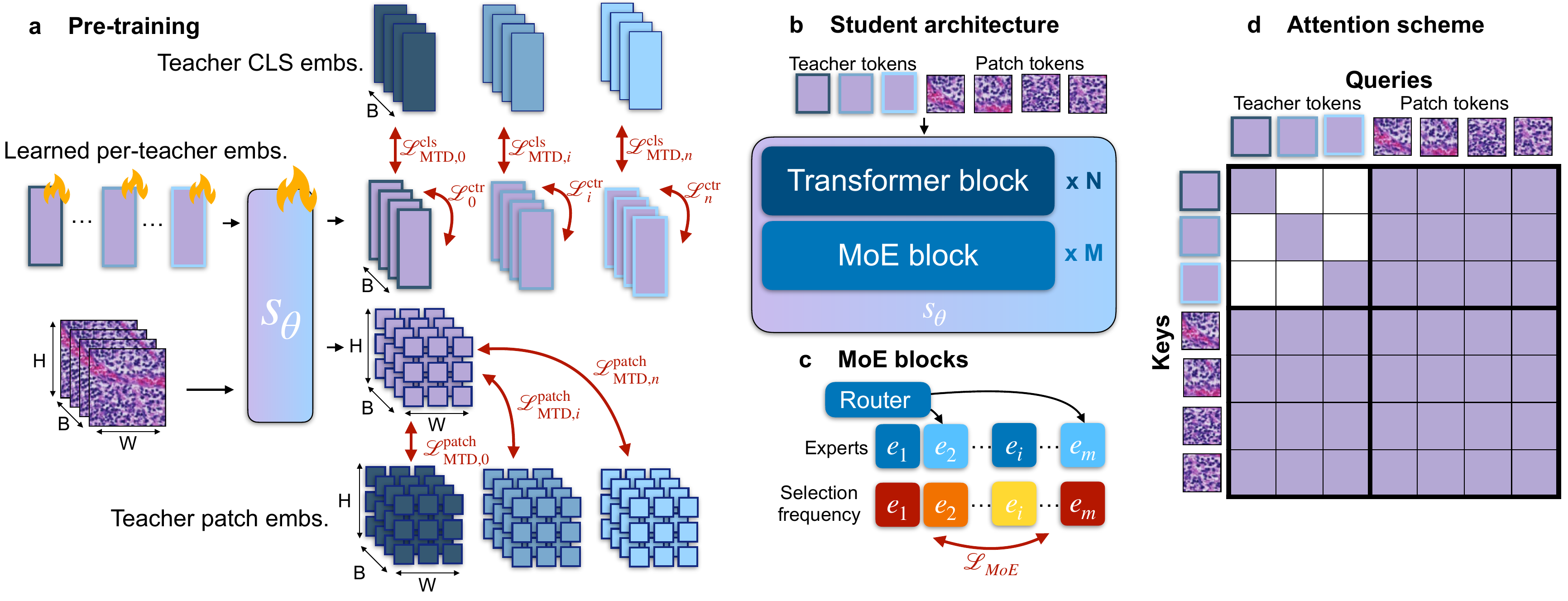}
  \caption{\textbf{Method overview}. \textbf{(a)} We train teacher-specific embeddings along with the student model with a token loss ($\mathcal{L}^{\mathrm{cls}}_{\mathrm{MTD},i}$ for each teacher $T_i$), a dense patch loss ($\mathcal{L}^{\mathrm{patch}}_{\mathrm{MTD},i}$) and a contrastive loss ($\mathcal{L}^{\mathrm{ctr}}_i$). \textbf{(b)} The student architecture is composed of a sequence of transformer and Mixture-of-Experts (MoE) layers. \textbf{(c)} Each MoE layer is based on an expert-selection mechanism where top-2 experts are picked conditioned on the input. An additional regularization loss ($\mathcal{L}_{\mathrm{MoE}}$) ensures a diversity in the selection of the experts. \textbf{(d)} All attention operations in the student follow a structured masking. Each teacher token attends to itself and to all patch tokens, but not to other teacher tokens, while patch tokens attend to all tokens.}   \label{fig:method}
\end{figure}
 
Given a training dataset $\mathcal{D} = \{d_j\}_{j=1...N}$ composed of $N$ samples, the vanilla MTD objective is,
\begin{equation}
\hat{\theta} = \arg \min_{\theta} \sum_{i=1}^{M} \sum_{j=1}^{N} \mathcal{L}^{\mathrm{cls}}_{\mathrm{MTD},i}(\theta, \psi_i, w_i, d_j)
  + \mathcal{L}^{\mathrm{patch}}_{\mathrm{MTD},i}(\theta, \psi_i, \bar{w}_i, d_j),
\end{equation}

where losses are themselves computed as follows,

\begin{equation}
\begin{split}
\mathcal{L}^{\mathrm{cls}}_{\mathrm{MTD},i}(\theta, \psi_i, w_i, d_j)
&=
\Phi\!\left[
p^{\mathrm{cls}}_{w_i}\!\left(s_{\theta}(d_j; \theta)\mathds{1}_{cls}; w_i\right),\
t_{\psi_i}(d_j;\psi_i) \mathds{1}_{cls}
\right], \\
\mathcal{L}^{\mathrm{patch}}_{\mathrm{MTD},i}(\theta, \psi_i, \bar{w}_i, d_j)
&=
\Phi\!\left[
p^{\mathrm{patch}}_{\bar{w}_i}\!\left(s_{\theta}(d_j; \theta)\mathds{1}_{patch}; \bar{w}_i\right),\
t_{\psi_i}(d_j;\psi_i) \mathds{1}_{patch}
\right],
\end{split}
\end{equation}

with $\mathds{1}_{cls}$ and $\mathds{1}_{patch}$ being indicator functions respectively indexing only CLS and patch tokens. $\Phi(\mathbf{x},\mathbf{y})= [1-\cos(\mathbf{x},\mathbf{y})] + \lambda_{L_1} \text{smooth}_{L_1}(\mathbf{x}-\mathbf{y})$ combines a cosine and smooth L1 losses, as done in previous work~\cite{Ranzinger_2024_CVPR, ranzinger2024phisdistributionbalancinglabelfree, shang2024theia, sariyildiz2025dune, sariyildiz2024unic, ma2024towards, grashei2025pathryoshka}.

\subsection{Teacher tokens and mixture-of-experts blocks}

A central challenge in multi-teacher distillation is that teachers might capture different information. A single shared CLS token must simultaneously compress all of these into one representation, which becomes increasingly bottlenecked as $M$ grows
(Figure~\ref{fig:teaser}b). We address this issue with two complementary design choices: teacher-specific tokens with a structured attention mask, and Mixture-of-Experts (MoE) layers. An overview of the pre-training scheme is presented in Figure~\ref{fig:method}a.

\myparagraph{Teacher-specific tokens}
In vanilla MTD, all $M$ teachers must be reconstructed from a single shared
CLS token $\mathbf{z}^{\mathrm{cls}}$, forcing it to simultaneously compress
potentially different visual abstractions into one vector. We therefore abandon the single-token design and instead assign one independent token per teacher.
Formally, we maintain a learnable token matrix $\mathbf{C} \in \mathbb{R}^{M \times D}$ and a corresponding positional embedding matrix $\mathbf{E}^{\mathrm{teach}} \in \mathbb{R}^{M \times D}$. For a given input and a set of $M$ active teachers, we prepend their $M$ specific tokens to the $N_p$ patch tokens, forming a sequence of length $M + N_p$:
\begin{equation}
  \mathbf{X} = \left[(\mathbf{C} + \mathbf{E}^{\mathrm{teach}})\,,\;
                      \mathbf{Z}^{\mathrm{patch}}\right]
  \in \mathbb{R}^{(M + N_p) \times D}.
\end{equation}
Each teacher-specific token can thus learn to selectively aggregate
the patch-level information most relevant to its assigned teacher.

\myparagraph{Structured attention masking} To encourage the learning of a specific aggregation of patch token embeddings from each teacher token, we apply in all the layers the following block-structured attention mask $\mathbf{A} \in \{0,1\}^{(M+N_p) \times (M+N_p)}$:

\begin{equation}
  \mathbf{A} =
  \left(
  \begin{array}{cc}
    {\color{BrickRed}\mathbf{I}_M} & {\color{RoyalBlue}\mathbf{1}_{M \times N_p}} \\[4pt]
    {\color{RoyalBlue}\mathbf{1}_{N_p \times M}} & {\color{OliveGreen}\mathbf{1}_{N_p \times N_p}}
  \end{array}
  \right),
  \label{eq:attn_mask}
\end{equation}
where a value of $1$ indicates that the query (row) can attend to the key
(column), and $0$ indicates masking.
The three colored blocks encode distinct attention roles:
(i) {\color{BrickRed}$\mathbf{I}_M$}: each token attends only to itself among teacher tokens;
(ii) {\color{RoyalBlue}$\mathbf{1}_{M \times N_p}$ and $\mathbf{1}_{N_p \times M}$}: every teacher-specific token attends to all patch tokens and vice versa, enabling full spatial aggregation per teacher;
(iii) {\color{OliveGreen}$\mathbf{1}_{N_p \times N_p}$} patch tokens attend freely among themselves, preserving standard self-attention over image content.

The masking of cross-teacher attention (Figure~\ref{fig:method}d) is an important design choice. By enforcing ${\color{BrickRed}\mathbf{I}_M}$ in the teacher-to-teacher block, we guarantee that each teacher token is exclusively informed by the patch tokens processed rather than by the already-aggregated views of other teachers. The objective here is to obtain a set of $M$ teacher-specialized representations that share computation at the patch level but diverge at the aggregation level.

\myparagraph{MoE blocks} As a complementary mechanism, the last transformer blocks replace the standard dense feed-forward networks with Mixture-of-Experts (MoE) layers, increasing the total model capacity while keeping inference cost constant. Again, to provide more flexibility in the extraction of image representations, for each token, a MoE router adaptively selects a set of relevant experts, i.e., feed-forward networks (Figure~\ref{fig:method}b, c). To encourage balanced expert utilization, we apply the Switch Transformer load-balancing loss~\cite{fedus2022switch}:
\begin{equation}
  \mathcal{L}_{\mathrm{MoE}} = E \cdot \sum_{e=1}^{E} f_e \cdot \bar{p}_e,
\end{equation}
where $f_e$ is the fraction of tokens routed to expert $e$ and $\bar{p}_e$
is the mean routing probability assigned to expert $e$ over the batch.
This loss is summed across all MoE blocks and added to the total objective
with coefficient $\alpha_{\mathrm{MoE}}{=}0.05$.

\subsection{Training objective}

\myparagraph{Per-teacher loss}
The total per-teacher loss combines CLS-level and patch-level cosine losses
with an InfoNCE contrastive term:
\begin{equation}
  \mathcal{L}_i
  = \mathcal{L}^{\mathrm{cls}}_{\mathrm{MTD},i}
  + \mathcal{L}^{\mathrm{patch}}_{\mathrm{MTD},i}
  + \lambda_{\mathrm{ctr}}\,\mathcal{L}^{\mathrm{ctr}}_i,
\end{equation}
with $\lambda_{\mathrm{ctr}}{=}0.1$.

\myparagraph{InfoNCE contrastive loss}
The contrastive term $\mathcal{L}^{\mathrm{ctr}}_i$ is a symmetric InfoNCE loss~\cite{oord2018representation} computed between the projected student teacher-specific embeddings and the corresponding ground-truth teacher CLS embeddings within a batch of size $B$. Let $\tilde{\mathbf{z}}^{(b)}_i = \ell_2\!\left(p^{\mathrm{cls}}_i(s_\theta(d_b))\right)$ and $\tilde{\mathbf{y}}^{(b)}_i = \ell_2\!\left(\mathbf{y}^{\mathrm{cls}}_i(d_b)\right)$ denote $\ell_2$-normalized student and teacher embeddings for sample $b$, 
where $s_\theta(d_b)$ is the student teacher-specific token for teacher $i$, 
$p^{\mathrm{cls}}_i$ the corresponding linear projection head, and 
$\mathbf{y}^{\mathrm{cls}}_i(d_b)$ the CLS embedding of teacher $i$.
The pairwise logit matrix $\mathbf{S}_i \in \mathbb{R}^{B \times B}$ is defined as
\begin{equation}
  [\mathbf{S}_i]_{ab} = \frac{\tilde{\mathbf{z}}^{(a)\top}_i\,\tilde{\mathbf{y}}^{(b)}_i}{\tau},
\end{equation}
with temperature $\tau{=}0.1$.
The InfoNCE loss is then the average of two cross-entropy terms that treat the
diagonal entries as the positives:
\begin{equation}
  \mathcal{L}^{\mathrm{ctr}}_i
  = \frac{1}{2}\left[
      -\frac{1}{B}\sum_{b=1}^{B}\log\frac{e^{[\mathbf{S}_i]_{bb}}}{\sum_{b'=1}^{B}e^{[\mathbf{S}_i]_{bb'}}}
      \;-\;
      \frac{1}{B}\sum_{b=1}^{B}\log\frac{e^{[\mathbf{S}_i]_{bb}}}{\sum_{a=1}^{B}e^{[\mathbf{S}_i]_{ab}}}
    \right].
  \label{eq:infonce}
\end{equation}
This batch-level discriminative signal complements the pointwise cosine
objective by penalizing embeddings that collapse to similar representations
across different samples.

\myparagraph{Final objective}
The full training loss sums over all $M$ teachers and incorporates the MoE loss:
\begin{equation}
  \mathcal{L}
  = \frac{1}{M}\sum_{i=1}^{M} \mathcal{L}_i
  + \alpha_{\mathrm{MoE}}\,\mathcal{L}_{\mathrm{MoE}}.
\end{equation}

\subsection{Downstream aggregation}

At downstream time, the student produces $M$ teacher-specific tokens which must be aggregated into a single representation. We use a simple mean-pooling operator which, as supported by our experimental results (Section~\ref{sec:experimental_results}),  leads to high performance. This might be explained by the disentanglement of teacher representations as studied in Figure~\ref{fig:cls_specialization}b. We also briefly experiment with learning a set of teacher-specific weights along with a MIL aggregator (Table~\ref{tab:bracs_slide}) to perform a weighted mean-pooling.

\section{Experimental results}
\label{sec:experimental_results}
\subsection{Experimental Setup}

\myparagraph{Pre-training dataset}
All distillation experiments use a pre-training dataset of $10$M tiles extracted
from $10$K whole-slide images sourced from The Cancer Genome Atlas (TCGA),
spanning a wide range of tissue types and organ systems, including lung, breast,
colon, kidney, and brain.

\myparagraph{Teacher models}
We consider a heterogeneous set of $M{=}10$ frozen teacher encoders covering both
histopathology-specific and general-purpose vision backbones, and both vision-only
and vision-language pretraining objectives.
The pool consists of:
H-optimus-1~\cite{hoptimus1},
Virchow2~\cite{zimmermann2024virchow2},
UNI2-h~\cite{chen2024towards},
Prov-GigaPath~\cite{xu2024whole},
Kaiko-ViT-B/8~\cite{aben2024towards},
CONCH~1.5~\cite{lu2024visual, ding2024multimodal},
Hibou-L~\cite{nechaev2024hibou},
H0-mini~\cite{filiot2025distilling},
KEEP~\cite{zhou2024knowledge},
and DINOv3-ViT-L/16~\cite{simeoni2025dinov3}.
These teachers differ in model capacity (from $86$M to $1.1$B parameters),
embedding dimensionality (from $384$ to $1{,}536$), and patch grid resolution
($7{\times}7$ to $28{\times}28$). All teachers are kept frozen throughout
student training.

\myparagraph{Concurrent distillation methods} We consider two close concurrent methods: H0-mini~\cite{filiot2025distilling}, which is part of our pool of teachers and is trained with single-teacher distillation, and GPFM~\cite{ma2024towards} trained with multi-teacher distillation. Checkpoints for both models are publicly released allowing us to compare them against OMNI. We do not compare with TICON~\cite{belagali2025ticon} which is a slide-level encoder and is thus not directly comparable, or with Pathryoshka~\cite{grashei2025pathryoshka} as no public code or pre-trained weights are currently publicly available. H0-mini is trained with a supervision coming from H-Optimus-0~\cite{hoptimus0}, and GPFM from UNI~\cite{chen2024towards}, Prov-GigaPath~\cite{xu2024whole} and Phikon~\cite{filiot2023scaling}. Both of them are trained on a set of publicly available slides (H0-mini: $6k$ WSIs, $43M$ tiles; GPFM: $72k$ WSIs, $190M$ tiles). We thus train our model on a more diverse set of teachers but from less data ($10k$ WSIs, $10M$ tiles). Despite these discrepancies, we believe that comparing OMNI to both methods is relevant as they are the current state-of-the-art open-source distilled models that can be used in clinical applications.

\myparagraph{Student model and MoE configuration}
We train a family of four student models of increasing capacity: OMNI-T ($7.3$M), OMNI-S ($28.7$M), OMNI-B ($114$M), and OMNI-L ($330$M). For each, the last $K$ transformer blocks use MoE layers with $E{=}5$ experts and top-$k{=}2$ routing ($K{=}3$ for T/S/B, $K{=}5$ for L), and the model outputs $M{=}10$ independent per-teacher CLS tokens. Full architecture details are provided in Appendix~\ref{sec:implem_details}.

\myparagraph{Tile-level evaluation}
Tile-level representations are evaluated using the \textbf{THUNDER}
benchmark~\cite{marza2025thunder}, which covers $13$ datasets spanning
diverse tissue types, cancer subtypes, and staining protocols:
BACH~\cite{aresta2019bach} (breast tissue subtypes),
BRACS~\cite{brancati2022bracs} (breast carcinoma subtypes),
BreakHis~\cite{spanhol2015dataset} (breast tumor magnification),
CCRCC~\cite{brummer2023computational} (clear-cell renal carcinoma),
CRC~\cite{kather2018100} (colorectal cancer tissue types),
ESCA~\cite{tolkach2023artificial} (esophageal carcinoma),
MHIST~\cite{10.1007/978-3-030-77211-6_2} (colorectal polyp grading),
PCam~\cite{veeling2018rotation} (lymph node metastasis detection),
WILDS~\cite{koh2021wilds} (multi-site breast cancer),
and four SPIDER sub-tasks~\cite{nechaev2025spider} covering bone marrow cytology
(SPIDER-B, SPIDER-C, SPIDER-S, SPIDER-T). Importantly, as we train our model on a subset of TCGA, we have removed the three TCGA-derived datasets from THUNDER to avoid data leakage that could advantage our method in empirical comparisons. 
We evaluate linear probing (LP), reporting macro-averaged F1 score
(Table~\ref{tab:lp_tile_slide}). We additionally evaluate adversarial robustness under a PGD attack ($\varepsilon{=}1.5{\times}10^{-3}$), reporting the absolute F1 drop relative to clean F1-score, as well as calibration via Expected Calibration Error (ECE) (Table~\ref{tab:robustness}). KNN and few-shot classification results are provided in the supplementary material (Table~\ref{tab:knn_fewshot}).

\myparagraph{Slide-level evaluation}
For slide-level evaluation we use the \textbf{PathoBench}
benchmark~\cite{zhang2025accelerating} on CPTAC cohorts~\cite{edwards2015cptac},
covering $7$ cancer types and $25$ tasks in total:
BRCA ($3$ tasks), CCRCC ($2$), COAD ($7$), GBM ($3$), LSCC ($4$),
LUAD ($4$), and UCEC ($2$).
Tasks include subtype classification and molecular marker prediction.
Following the PathoBench protocol, we report balanced accuracy (BACC), and we evaluate the CPTAC tasks using a 50-fold train/test split, while employing 5-fold cross-validation for all other WSI datasets. For all splits, we ensure there is no data leakage between the training and test sets.
Slide-level tile features are extracted using the TRIDENT
pipeline~\cite{zhang2025accelerating} and aggregated via mean pooling, followed by a linear
classifier (Table~\ref{tab:lp_tile_slide}).

Further hyperparameters and implementation details are
provided in Appendix (Sections~\ref{sec:patch_loss_details},~\ref{sec:implem_details},~\ref{sec:downstream_details}).

\myparagraph{No data leakage caused by an increased number of teachers} According to the publicly-available information, no test dataset from our downstream evaluations is included in the training set of the teacher foundation models. We base our study on the reported training datasets in Table S1 (appendix) from~\cite{marza2025thunder}, which is partially derived from Table 1 in~\cite{li2025survey}. As can be seen in these tables, our considered teachers were trained from private sources, \textit{TCGA}, \textit{GTEx}, \textit{Quilt1M}, \textit{OpenPath} and natural-domain datasets, which, to the best of our knowledge, do not overlap with our test datasets.

\begin{table}[t]
\centering
\caption{%
    \label{tab:lp_tile_slide}
    Linear probing performance at tile level (F1\,\%, 13 datasets),
    slide level (Bacc\,\%, 7 CPTAC datasets / 25 tasks),
    and their mean.
    Mean rank is computed per dataset then averaged. * indicates significance of paired Wilcoxon signed-rank test vs.\ OMNI-L:
$^* p < 0.05$; $^{**} p < 0.01$. 
    \textbf{Bold}: best; \underline{underline}: second best per column.}
\resizebox{\textwidth}{!}{
  \begin{tabular}{llllcc|cc|cc}
    \toprule
    & & & & \multicolumn{2}{c|}{\textbf{Tile-level}} & 
    \multicolumn{2}{c|}{\textbf{Slide-level}} & \multicolumn{2}{c}{\textbf{Mean}} \\
    \cmidrule(lr){5-6} \cmidrule(lr){7-8} \cmidrule(lr){9-10}
    \textbf{Model} & \textbf{Params.} & \textbf{Dim} & \textbf{GFLOPs} & \textbf{F1\,(\%)} & \textbf{Rank\,$(\downarrow)$} & \textbf{Bacc\,(\%)} & \textbf{Rank\,$(\downarrow)$} & \textbf{Score} & \textbf{Rank\,$(\downarrow)$} \\
    & & & & {\small\textcolor{gray}{13 ds.}} & 
    {\small\textcolor{gray}{13 ds.}} & {\small\textcolor{gray}{25 tasks}} & {\small\textcolor{gray}{7 datasets}} &  &  \\
    \midrule
     \multicolumn{10}{l}{\textit{\small Foundation models}} \\
     \quad UNI2-h~\cite{chen2024towards} & 681M & 1536 & 180.4 & \underline{87.4}\phantom{$^{**}$} & 4.6 & 61.5$^{**}$ & 9.0 & 74.5 & 6.8 \\
    \quad Virchow2~\cite{zimmermann2024virchow2} & 631M & 1280 & 164.6 & 85.9\phantom{$^{**}$} & 6.6 & 61.5$^*$\phantom{$^{*}$} & 8.3 & 73.7 & 7.4 \\
    \quad H-optimus-1~\cite{hoptimus1} & 1.1B & 1536 & 296.0 & 86.3$^*$\phantom{$^{*}$} & 6.2 & 62.5\phantom{$^{**}$} & 4.7 & 74.4 & 5.4 \\
    \quad KEEP~\cite{zhou2024knowledge} & 414M & 1024 & 59.7 & 84.9$^*$\phantom{$^{*}$} & 9.5 & 60.7$^{**}$ & 11.0 & 72.8 & 10.2 \\
    \quad Prov-GigaPath~\cite{xu2024whole} & 1.1B & 1536 & 223.4 & 84.2$^{**}$ & 9.3 & 61.0$^{**}$ & 10.7 & 72.6 & 10.0 \\
    \quad CONCH 1.5~\cite{lu2024visual,ding2024multimodal} & 307M & 1024 & 79.54 & 84.3$^{**}$ & 11.0 & 61.5$^*$\phantom{$^{*}$} & 9.4 & 72.9 & 10.2 \\
    \quad Hibou-B~\cite{nechaev2024hibou} & 86M & 768 & 21.97 & 82.5$^{**}$ & 10.7 & 60.3$^{**}$ & 11.0 & 71.4 & 10.8 \\
    \quad Kaiko-ViT-B/8~\cite{aben2024towards} & 86M & 768 & 66.86 & 82.0$^{**}$ & 13.2 & 58.9$^{**}$ & 13.9 & 70.5 & 13.6 \\
    \cmidrule(lr){1-10}
    \multicolumn{10}{l}{\textit{\small Distilled models}} \\
    \quad GPFM~\cite{ma2024towards} & 303M & 1024 & 81.08& 83.2$^{**}$ & 9.5 & 60.9$^{**}$ & 9.1 & 72.0 & 9.3 \\
    \quad H0-mini~\cite{filiot2025distilling} & 86M & 768 & 22.3 & 84.9$^{**}$ & 8.3 & 61.5$^{**}$ & 9.4 & 73.2 & 8.9 \\
    \cmidrule(lr){1-10}
    \multicolumn{10}{l}{\textit{\small Vanilla MTD}} \\
    \quad ViT-B single CLS & 86M & 768 & 21.97 & 85.4$^*$\phantom{$^{*}$} & 5.9 & 61.8$^{**}$ & 8.7 & 73.6 & 7.3 \\
    \midrule
    \multicolumn{10}{l}{\textit{\small Ours}} \\
    \quad \textbf{OMNI-T} & 7.3M & 192 & \textbf{1.39} & 83.9\phantom{$^{**}$} & 11.0 & \underline{63.2}\phantom{$^{**}$} & \underline{4.0} & 73.6 & 7.5 \\
    \quad \textbf{OMNI-S} & 28.7M & 384 & \underline{5.48} & 85.3\phantom{$^{**}$} & 7.6 & 62.9\phantom{$^{**}$} & 4.1 & 74.1 & 5.8 \\
    \quad \textbf{OMNI-B} & 114M & 768 & 21.8 & 87.1\phantom{$^{**}$} & \underline{4.0} & \underline{63.2}\phantom{$^{**}$} & 4.4 & \underline{75.2} & \underline{4.2} \\
    \quad \textbf{OMNI-L} & 330M & 1024 & 67.45 & \textbf{87.8}\phantom{$^{**}$} & \textbf{2.5} & \textbf{63.5}\phantom{$^{**}$} & \textbf{2.1} & \textbf{75.7} & \textbf{2.3} \\
    \bottomrule
  \end{tabular}
}
\end{table}

\begin{figure}[t]
  \begin{subfigure}[t]{0.32\textwidth}
        \centering
        \includegraphics[width=\linewidth]{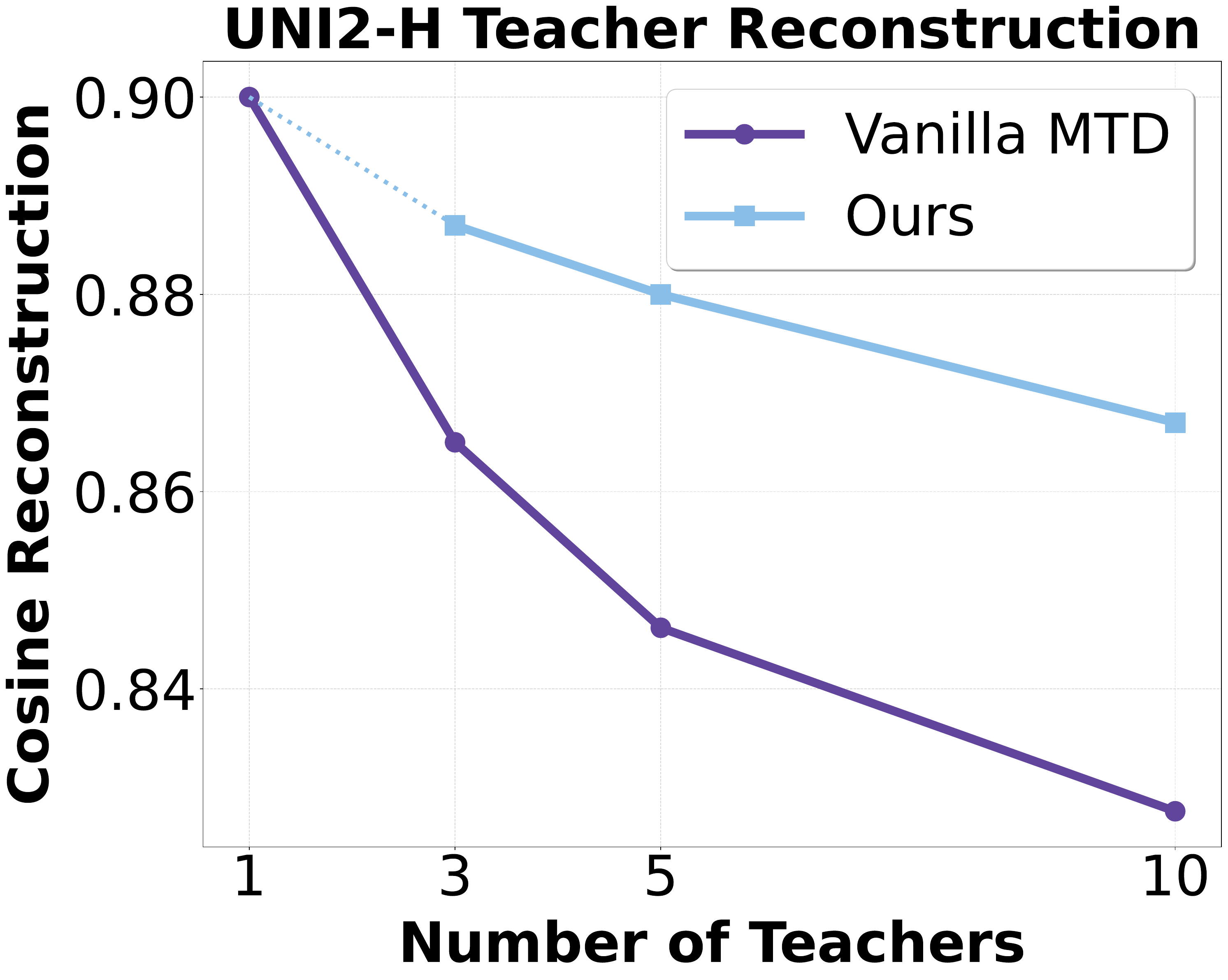}
        \caption{}
    \end{subfigure}
    ~ 
    \begin{subfigure}[t]{0.32\textwidth}
        \centering
        \includegraphics[width=\linewidth]{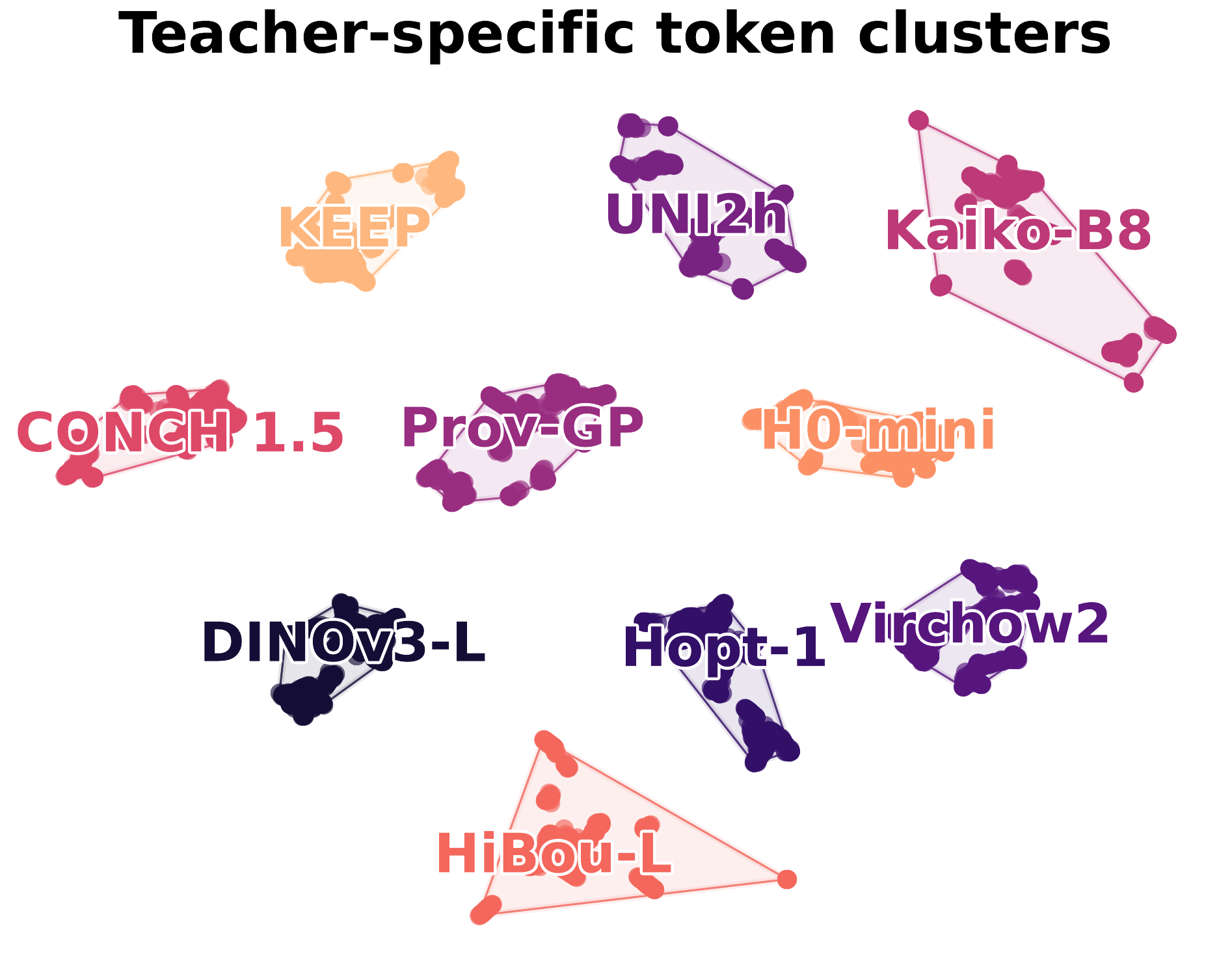}
        \caption{}
    \end{subfigure}
    ~ 
    \begin{subfigure}[t]{0.32\textwidth}
        \centering
        \includegraphics[width=\linewidth]{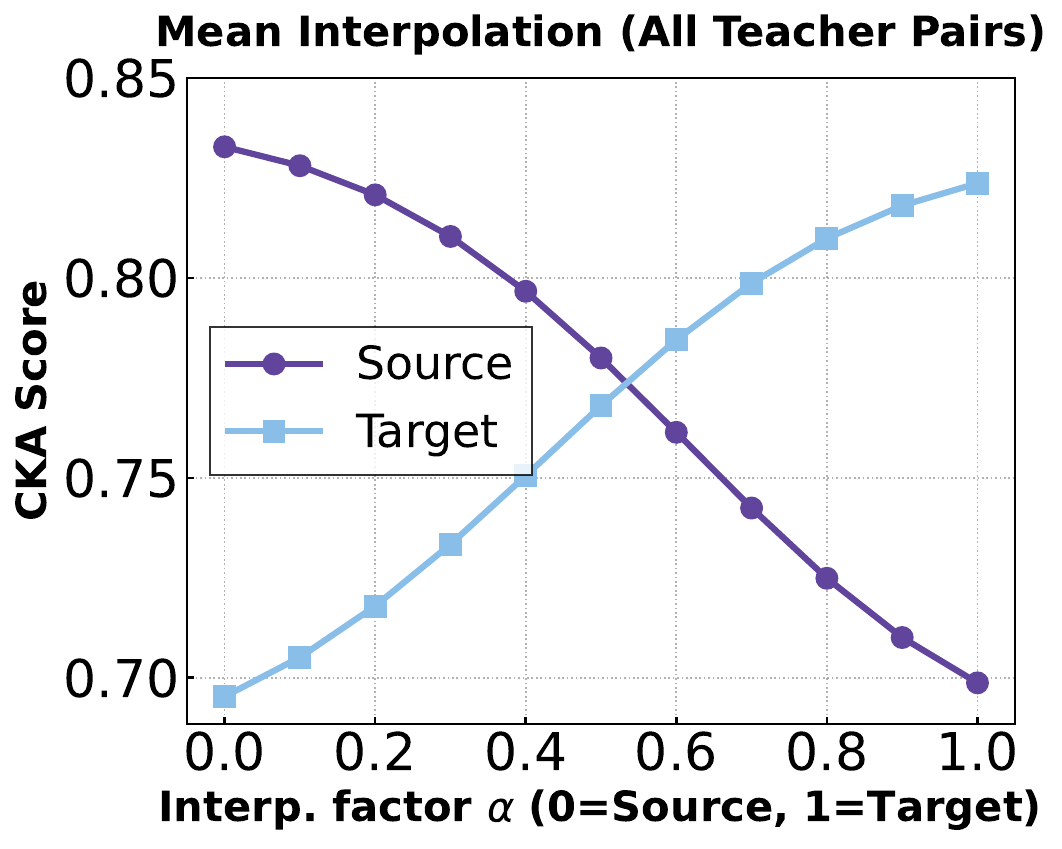}
        \caption{}
    \end{subfigure}
  \caption{\textbf{Analysis of teacher representations}. \textbf{(a)} Cosine reconstruction quality of UNI2-H as the number of teachers increases. As teachers compete for a single shared token, vanilla MTD reconstruction quality degrades significantly, whereas our approach is substantially more robust to the growing number of teachers. \textbf{(b)} UMAP of the $M{=}10$ teacher-specific tokens from OMNI-B on 3$k$ images.
    Well-separated clusters confirm each token specializes toward its assigned 
    teacher. \textbf{(c)} Smooth latent interpolation between two teacher-specific tokens (averaged over all teacher pairs), showing a consistent X-shaped crossover pattern as the conditioning vector is linearly interpolated.}
  \label{fig:cls_specialization}
\end{figure}

\begin{table}[t]
  \centering
  \setlength\tabcolsep{4pt}
  % ---------------- Left: BRACS ----------------
  \begin{minipage}[t]{0.47\linewidth}
    \centering
    \caption{ABMIL slide-level results on BRACS (7-class subtyping). Balanced
    accuracy (\%), mean\,$\pm$\,std across 5 folds. $^\dagger$Weighted mean pooling, where a scalar weight per teacher token is
    learned jointly with the ABMIL aggregator on the downstream task.}
    \label{tab:bracs_slide}
    \begin{tabular}{lc}
      \toprule
      \textbf{Model} & \textbf{BACC\,(\%)} $(\uparrow)$ \\
      \midrule
      UNI2-h~\cite{chen2024towards}                    & $47.5 \pm 1.7$ \\
      Virchow2~\cite{zimmermann2024virchow2}           & $44.1 \pm 0.8$ \\
      H-optimus-1~\cite{hoptimus1}                     & $41.6 \pm 2.3$ \\
      KEEP~\cite{zhou2024knowledge}                    & $46.9 \pm 1.7$ \\
      H0-mini~\cite{filiot2025distilling}              & $44.5 \pm 2.1$ \\
      Prov-GigaPath~\cite{xu2024whole}                 & $41.5 \pm 1.6$ \\
      CONCH~1.5~\cite{lu2024visual,ding2024multimodal} & $46.2 \pm 1.0$ \\
      GPFM~\cite{ma2024towards}                        & $46.2 \pm 1.0$ \\
      Vanilla MTD                                      & $43.5 \pm 2.2$ \\
      \midrule
      \textbf{OMNI-T (ours)}           & $\underline{49.2 \pm 1.6}$ \\
      \textbf{OMNI-T$^\dagger$ (ours)} & $\mathbf{49.9 \pm 2.0}$ \\
      \textbf{OMNI-S (ours)}           & $47.2 \pm 0.9$ \\
      \textbf{OMNI-S$^\dagger$ (ours)} & $46.0 \pm 1.1$ \\
      \textbf{OMNI-B (ours)}           & $47.2 \pm 2.7$ \\
      \textbf{OMNI-B$^\dagger$ (ours)} & $\mathbf{49.9 \pm 3.0}$ \\
      \textbf{OMNI-L (ours)}           & $45.2 \pm 1.2$ \\
      \textbf{OMNI-L$^\dagger$ (ours)} & $46.3 \pm 1.2$ \\
      \bottomrule
    \end{tabular}
  \end{minipage}
  \hfill
  % ---------------- Right: Robustness + UNIC/DUNE ----------------
  \begin{minipage}[t]{0.50\linewidth}
    \centering
    \caption{Adversarial robustness (mean F1 drop, \%) under PGD
    ($\varepsilon{=}1.5{\times}10^{-3}$) and calibration (mean ECE, \%), averaged across
    13 datasets.}
    \label{tab:robustness}
    \begin{tabular}{lcc}
      \toprule
      \textbf{Model} & \textbf{F1 drop} $(\downarrow)$ & \textbf{ECE} $(\downarrow)$ \\
      \midrule
      UNI2-h~\cite{chen2024towards}                    & 30.4 & 4.0 \\
      Virchow2~\cite{zimmermann2024virchow2}           & 29.4 & 3.8 \\
      H-optimus-1~\cite{hoptimus1}                     & 58.1 & 4.0 \\
      KEEP~\cite{zhou2024knowledge}                    & 43.9 & 4.4 \\
      H0-mini~\cite{filiot2025distilling}              & 33.9 & 4.3 \\
      Kaiko-ViT-B/8~\cite{aben2024towards}             & 39.4 & 3.5 \\
      Prov-GigaPath~\cite{xu2024whole}                 & 40.8 & 3.7 \\
      CONCH~1.5~\cite{lu2024visual,ding2024multimodal} & 77.1 & 4.7 \\
      \midrule
      \textbf{OMNI-T (ours)} & 31.6 & \underline{3.4} \\
      \textbf{OMNI-S (ours)} & 22.3 & 3.9 \\
      \textbf{OMNI-B (ours)} & \underline{20.1} & 3.7 \\
      \textbf{OMNI-L (ours)} & \textbf{16.1} & \textbf{2.8} \\
      \bottomrule
    \end{tabular}
    \caption{Comparison to GPFM~\cite{ma2024towards}, with both models trained
    on the same 3 teachers. F1 (\%) averaged across 13 THUNDER tasks.}
    \label{tab:gpfm_comparison}
    \begin{tabular}{lc}
      \toprule
      \textbf{Model} & \textbf{F1\,(\%)} $(\uparrow)$ \\
      \midrule
      GPFM~\cite{ma2024towards} & 83.2 \\
      \textbf{OMNI-L (ours)}    & \textbf{85.5} \\
      \bottomrule
    \end{tabular}
  \end{minipage}
   \vspace{-0.5em}

\end{table}

\subsection{Downstream performance}

\myparagraph{Linear probing benchmarking} Table~\ref{tab:lp_tile_slide} reports linear probing performance across tile- and slide-level benchmarks. The full OMNI family demonstrates a consistent scaling trend: even OMNI-T, with only $7.3$M parameters and $1.39$ GFLOPs, achieves competitive slide-level performance ($63.2$ Bacc), while OMNI-S already surpasses most distilled baselines at a fraction of their computational cost. OMNI-B and OMNI-L push further, by being ranked second and first respectively, while being up to $10\times$ more efficient in GFLOPs than the best teacher models. Notably, OMNI-L achieves the best tile-level F1 ($87.8\%$) and slide-level Bacc ($63.5\%$), outperforming all teachers including UNI2-h ($681$M) and H-optimus-1 ($1.1$B) despite being significantly smaller. Both OMNI-B and OMNI-L also substantially outperform the concurrent multi-teacher distillation baseline GPFM across all metrics, with GPFM ranking below all OMNI variants despite using $303$M parameters. To assess statistical significance, we performed paired Wilcoxon signed-rank tests between OMNI-L and all foundation models: OMNI-L outperforms significantly the large majority of teachers at both tile and slide level ($^* p < 0.05$; $^{**} p < 0.01$). Additional results on the THUNDER benchmark for the knn and few-shot classification tasks are presented in the appendix (Section~\ref{sec:additional_res}, Table~\ref{tab:knn_fewshot}).

\myparagraph{Gains when combined with a stronger aggregator} In addition to lightweight mean-pooling, we also evaluate the performance of our method with ABMIL~\cite{ilse2018attention} aggregation on the BRACS dataset. Table~\ref{tab:bracs_slide} shows that OMNI also performs well when coupled with ABMIL. We also experimented with optimizing the weights of all teacher-specific tokens. Instead of uniformly averaging teacher tokens, learning teacher-specific weights increases performance, leading to our method ranking first.

\myparagraph{Comparison with GPFM} We have compared our model to the publicly-released GPFM model in Table~\ref{tab:lp_tile_slide}. Additionally, we pre-train OMNI-L (to match the number of parameters of GPFM ViT-L) on the same set of 3 teachers used by GPFM (UNI, Prov-GigaPath, PHIKON). The only difference is thus about the used training dataset, but one should note that our dataset is a small subset of the one used to pre-train GPFM. As seen in Table~\ref{tab:gpfm_comparison}, when trained on the same teachers, a smaller dataset and with the same parameter size, we outperform the pre-trained GPFM model.

\begin{wraptable}[12]{r}{0.35\linewidth}
  \centering
  \small
  \setlength\tabcolsep{0.0pt}
         \caption{Main contributions of natural-domain MTD methods (1) UNIC and
  (2) DUNE, applied on top of vanilla MTD or OMNI-B. F1 averaged over 13 THUNDER
  tasks.}
  \label{tab:unic_dune_comp}
  \begin{tabular}{lc}
    \toprule
    \textbf{Model} & \textbf{F1\,(\%)} $(\uparrow)$ \\
    \midrule
    Vanilla MTD                                      & 85.4 \\
    \quad + Teacher drop.~\cite{sariyildiz2024unic} (1) & 85.6 \\
    \quad + Transf.\ decoder~\cite{sariyildiz2025dune} (2) & 85.9 \\
    \midrule
    \textbf{OMNI-B (ours)} & \textbf{87.1} \\
    \quad + (1) + (2)      & \underline{87.0} \\
    \bottomrule
  \end{tabular}
\end{wraptable}
\myparagraph{Comparison with natural-domain MTD methods} We consider two recent
state-of-the-art multi-teacher distillation approaches, namely
UNIC~\cite{sariyildiz2024unic} and DUNE~\cite{sariyildiz2025dune}, and apply
their main contributions that could be integrated into our framework and have
the potential to also improve the scaling abilities of the vanilla baseline. In
particular, we evaluate the gains coming from the teacher dropping strategy in
UNIC and the Transformer-based decoder architecture in DUNE.
Table~\ref{tab:unic_dune_comp} shows that while they improve the downstream
performance of the vanilla baseline, the gains are notably lower than the ones
brought by our method. Additionally, adding them on top of OMNI does not bring
further improvement, showing that such design choices are not needed along
with our method.

\myparagraph{Uncertainty and robustness} Table~\ref{tab:robustness} reports the
robustness to adversarial attacks and the calibration of linear probes trained
on top of embeddings from various foundation models, including ours, following
the THUNDER benchmark~\cite{marza2025thunder}. OMNI-L is both the most robust
to adversarial attacks, with an F1 drop of $16.1\%$ compared with $29.4\%$ for
the most robust teacher (Virchow2), and the best calibrated model, with an ECE
of $2.8\%$. OMNI-B and OMNI-S are also more robust than all teacher models.
This could be explained by our diverse teacher-specific representations which,
when fused, lead to a more robust representation.

\myparagraph{Ablation studies} Table~\ref{tab:ablation} ablates the importance of our proposed attention masking (\textit{OMNI-B} vs \textit{No attn. mask}), MoE blocks (\textit{OMNI-B} vs \textit{No MoE tail}), and infoNCE loss (\textit{OMNI-B} vs \textit{No InfoNCE loss}). When removing each contribution, performance drops compared with \textit{OMNI-B}, showing that all of them are necessary. All variants match or outperform on average \textit{Vanilla MTD} and the combination of all contributions reaches the highest performance.

\begin{table}[t]
  \centering
  \begin{minipage}[t]{0.74\linewidth}
    \centering
    \caption{Ablation study on the contribution of the structured attention,
    InfoNCE loss, and MoE blocks. A \ding{55} indicates that the corresponding
    component is \emph{removed} from the full OMNI-B model. Experiments are
    conducted on ViT-Base.}
    \label{tab:ablation}
    \resizebox{\linewidth}{!}{%
    \begin{tabular}{lcccccl}
      \toprule
      \textbf{Model}
        & \textbf{InfoNCE}
        & \textbf{MoE}
        & \textbf{Attn.\ mask}
        & \textbf{Tile} (F1)
        & \textbf{Slide} (Bacc)
        & \textbf{Average} \\
      \midrule
      OMNI-B (ours)    &           &            &            & \textbf{87.1} & \textbf{63.2} & \textbf{75.2} \\
      \midrule
      No InfoNCE loss  & \ding{55} &            &            & 86.6 ({\color{BrickRed}-0.5}) & 61.1 ({\color{BrickRed}-2.1}) & 73.9 ({\color{BrickRed}-1.3}) \\
      No MoE tail      &           & \ding{55}  &            & 86.6 ({\color{BrickRed}-0.5}) & 61.2 ({\color{BrickRed}-2.0}) & 73.9 ({\color{BrickRed}-1.3}) \\
      No attn.\ mask   &           &            & \ding{55}  & 86.2 ({\color{BrickRed}-0.9}) & 61.6 ({\color{BrickRed}-1.6}) & 73.9 ({\color{BrickRed}-1.3}) \\
      \midrule
      Vanilla MTD      & \ding{55} & \ding{55}  & \ding{55}  & 85.4 ({\color{BrickRed}-1.7}) & 61.8 ({\color{BrickRed}-1.4}) & 73.6 ({\color{BrickRed}-1.6}) \\
      \bottomrule
    \end{tabular}}
  \end{minipage}
  \hfill
  \begin{minipage}[t]{0.24\linewidth}
    \centering
    \caption{Effect of full isolation of teacher tokens. F1 (\%) averaged
    across 13 THUNDER tasks.}
    \label{tab:struct_att}
    {
    \small
    \begin{tabular}{lc}
      \toprule
      \textbf{Model} & \textbf{F1} \\
      \midrule
      OMNI-B        & \textbf{87.1} \\
      + full iso.   & 86.7 ({\color{BrickRed}-0.4}) \\
      \bottomrule
    \end{tabular}
    }
  \end{minipage}
\end{table}

\myparagraph{Structured attention masking} Our current design does not fully isolate teacher tokens from each other as information can be passed implicitly through multiple layers. However, we believe that such a formulation already serves the purpose of specializing different teacher tokens. Figure~\ref{fig:cls_specialization}b indeed shows that despite a lack of an explicit full isolation, it leads to disentangled teacher-specific representations. The usefulness of our attention mask formulation is empirically showcased in Table~\ref{tab:ablation} highlighting the drop in downstream performance if replacing it with vanilla self-attention between all tokens. We conduct an experiment where patch tokens can only attend to each other, this time having a full isolation. Table~\ref{tab:struct_att} reports the performance of such a baseline and, as can be seen, a stricter full isolation does not seem to bring further improvement and might not be necessary.

\subsection{Analysis of per-teacher representations}

\myparagraph{Teacher token specialization and scaling}
Figure~\ref{fig:cls_specialization}a shows that vanilla MTD suffers a steep degradation in reconstruction quality as the number of teachers grows, dropping from $0.90$ cosine similarity at $1$ teacher to $0.83$ at $10$. Our approach significantly mitigates this collapse, retaining $0.87$ at $10$ teachers. This demonstrates that the teacher-specific tokens with structured attention masking preserve individual reconstruction quality even under heavy multi-teacher pressure. The UMAP visualization in Figure~\ref{fig:cls_specialization}b further confirms that each teacher token specializes, with well-separated clusters across all $10$ teacher slots, rather than collapsing into a common average embedding.

\myparagraph{Latent space geometry}
Figure~\ref{fig:cls_specialization}c reveals a smooth and monotonic crossover when linearly interpolating two teacher tokens. It indeed shows the evolution of the average alignment (CKA similarity) between ground-truth teacher embeddings and the embedding resulting from the linear interpolation of two teacher-specific tokens as a function of their respective weights. As can be seen, the average CKA similarity to the source decreases steadily while similarity to the target increases, with a crossover near $\alpha \approx 0.55$. This indicates that the student model has learned a geometrically well-structured latent space in which teacher representations are continuously interpolable.

\section{Conclusion}
\label{sec:conclusion}
Multi-teacher distillation is a powerful framework to train efficient encoders from the supervision of diverse foundation models. In this paper, we show that vanilla MTD is not scaling well to an increasing number of teachers. We thus proposed a new per-teacher representation based on key design choices, namely learnable teacher tokens, a teacher-specific attention scheme, MoE layers and a per-teacher infoNCE contrastive loss. These contributions allowed us to reach state-of-the-art downstream performance and outperform a large set of foundation models, while being up to $10\times$ more efficient and with many fewer parameters. This led us to release a collection of efficient models called \textbf{OMNI} with different sizes, i.e., OMNI-T, OMNI-S, OMNI-B, OMNI-L. We believe that such lightweight and powerful vision encoders are well suited for clinical applications requiring high-quality feature extraction at a low compute cost.

\myparagraph{Limitations} Thanks to the complementarity of our teacher-specific representations, we rely on a simple mean-pooling aggregation mechanism. While we have explored optimizing a task-specific weighting of teachers on the BRACS dataset, this direction has yet to be studied more. Additionally, despite using the standard MTD patch loss, we did not explore the learning of teacher-specific dense representations. Both research directions are left for future work. Despite our empirical validation being composed of $39$ tasks spanning diverse public datasets, organs, magnifications and institutions, the clinical use of our model would require further validation.

\myparagraph{Broader impact} New foundation models have the potential to benefit positively clinical decisions and ultimately patient outcomes. However, such a clinical deployment should follow regulatory guidelines that come with careful validations and important considerations of ethical challenges.

\myparagraph{Acknowledgments} This work has been supported by the \textit{Agence Nationale de la Recherche} through \textit{ANR-23-IAHU-0002} (\textit{IHU PRISM}), \textit{ANR-21-CE45-0007}, \textit{ANR-23-CE45-0029}, \textit{ANR-23-IACL-0003} (\textit{DATAIA CLUSTER}) and the \textit{Health Data Hub} as part of the second edition of the \textit{France-Québec} call for projects \textit{Intelligence Artificielle en santé}. This work was performed using HPC resources from \textit{GENCI-IDRIS} (Grant \textit{2025-AD011015593R1}). This research was
also partially supported by NSF grants IIS-2123920, IIS-2212046.

\bibliographystyle{plainnat}
\bibliography{neurips_2026}

%%%%%%%%%%%%%%%%%%%%%%%%%%%%%%%%%%%%%%%%%%%%%%%%%%%%%%%%%%%%
\newpage
\appendix

\begin{center}
{\Large \textbf{Appendix}}
\end{center}

\section{Additional results}
\label{sec:additional_res}

\myparagraph{knn and few-shot classification} Table~\ref{tab:knn_fewshot}
reports knn and 16-shot few-shot classification results on the THUNDER
benchmark~\cite{marza2025thunder} across 13 datasets. OMNI-B and OMNI-L achieve
the best mean rank on knn, and OMNI-L ranks second on 16-shot few-shot
classification, behind UNI2-h. Both models remain significantly more efficient
than the top teacher models, with OMNI-B requiring about $8\times$ fewer
GFLOPs than UNI2-h.

\begin{figure}[!htbp]
  \centering
  \includegraphics[width=0.6\linewidth]{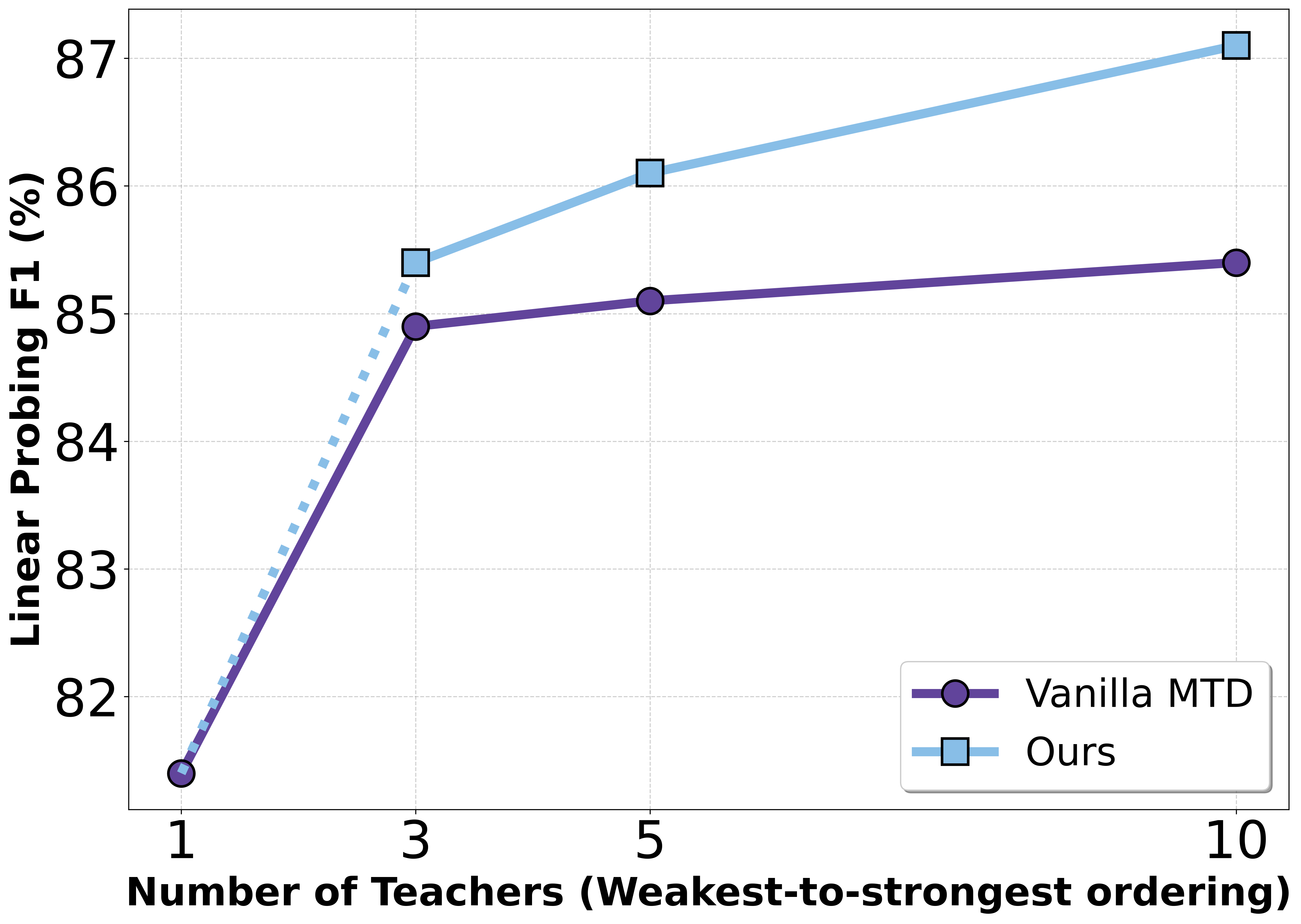}
  \caption{\textbf{Scaling with a reverse teacher ordering.} Tile-level linear
  probing F1 (\%) averaged across 13 THUNDER tasks as a function of the number
  of teachers, when adding teachers from weakest to strongest. Our method keeps
  improving as teachers are added, while vanilla MTD saturates.}
  \label{fig:lp_f1_reverse}
\end{figure}

\myparagraph{Robustness to the teacher ordering} In
Figure~\ref{fig:teaser}b, teachers are ranked by their individual performance
on the THUNDER benchmark and added from strongest to weakest. To verify that
the observed scaling trend does not depend on this ordering, we repeat the
experiment with the reverse order, adding teachers from weakest to strongest.
Results are reported in Figure~\ref{fig:lp_f1_reverse}. In this setting, both
methods benefit from the stronger teachers added at each step. However, vanilla
MTD quickly saturates, gaining only $0.5$ F1 points when going from $3$ to $10$
teachers. In contrast, our method keeps improving as more teachers are added,
and the gap between the two methods widens with the number of teachers, from
$0.5$ points with $3$ teachers to $1.7$ points with $10$ teachers. This
confirms that the better scaling of our method holds regardless of the order in
which teachers are integrated.

\begin{table}[h!]
  \centering
  \caption{%
    KNN and 16-shot few-shot classification results (F1\,\%, 13 datasets).
    Mean rank is computed per dataset then averaged.
    \textbf{Bold}: best; \underline{underline}: second best per column.}
  \label{tab:knn_fewshot}
  \begin{tabular}{llllcc|cc}
    \toprule
    & & & & \multicolumn{2}{c|}{\textbf{KNN}} & \multicolumn{2}{c}{\textbf{16-shot}} \\
    \cmidrule(lr){5-6} \cmidrule(lr){7-8}
    \textbf{Model} & \textbf{Params.} & \textbf{Dim} & \textbf{GFLOPs} & \textbf{F1\,(\%)} & \textbf{Rank\,$(\downarrow)$} & \textbf{F1\,(\%)} & \textbf{Rank\,$(\downarrow)$} \\
    & & & & {\small 13 ds.} & {\small 13 ds.} & {\small 13 ds.} & {\small 13 ds.} \\
    \midrule
    UNI2-h & 681M & 1536 & 180.4 & \textbf{85.0} & 4.7 & \textbf{81.9} & \textbf{3.4} \\
    Virchow2 & 631M & 1280 & 164.6 & 84.7 & 5.0 & 75.9 & 9.6 \\
    H-optimus-1 & 1.1B & 1536 & 296.0 & 83.9 & 5.3 & 78.3 & 7.1 \\
    KEEP & 414M & 1024 & 59.7 & 83.4 & 7.7 & 79.3 & 7.4 \\
    Prov-GigaPath & 1.1B & 1536 & 223.4 & 81.3 & 9.1 & 77.2 & 8.3 \\
    CONCH 1.5 & 307M & 1024 & 79.54 & 82.0 & 9.4 & 77.2 & 9.1 \\
    Kaiko-ViT-B/8 & 86M & 768 & 66.86 & 80.0 & 10.5 & 78.6 & 7.5 \\
    \cmidrule(lr){1-8}
    \multicolumn{8}{l}{\textit{\small Distilled models}} \\
    \quad GPFM & 303M & 1024 & 81.08 & 82.2 & 7.9 & 77.8 & 7.2 \\
    \quad H0-mini & 86M & 768 & 22.3 & 81.4 & 8.4 & 76.6 & 8.8 \\
    \midrule
    \textbf{OMNI-T (ours)} & 7.3M & 192 & 1.39 & 81.5 & 8.8 & 77.5 & 8.1 \\
    \textbf{OMNI-S (ours)} & 28.7M & 384 & 5.48 & 82.3 & 6.9 & 78.7 & 6.0 \\
    \textbf{OMNI-B (ours)} & 114M & 768 & 21.8 & 84.3 & \textbf{3.6} & 79.9 & 4.8 \\
    \textbf{OMNI-L (ours)} & 330M & 1024 & 67.45 & \underline{84.8} & \textbf{3.6} & \underline{80.7} & \underline{3.7} \\
    \bottomrule
  \end{tabular}
\end{table}

\myparagraph{Additional slide-level evaluation on IMP} To further increase the
diversity of our evaluation, we also benchmark our method on the IMP colorectal
cancer dataset~\cite{neto2024interpretable} from PathoBench, using an ABMIL
aggregator~\cite{ilse2018attention}. As the IMP dataset comprises around $5{,}000$
WSIs totaling more than $5$TB, embedding it with every model requires significant
compute. We thus restrict this evaluation to a subset of representative and
competitive baselines. As shown in Table~\ref{tab:imp_res}, OMNI-B outperforms
the vanilla MTD baseline and H0-mini, and matches the state-of-the-art teacher
foundation model UNI2-h while requiring about $8\times$ fewer GFLOPs.

\begin{table}[ht]
  \centering
  \caption{ABMIL slide-level classification on the IMP colorectal dataset
  (PathoBench). Balanced accuracy (\%) with 95\% CI and efficiency (GFLOPs).}
  \label{tab:imp_res}
  \begin{tabular}{lcc}
    \toprule
    Model & GFLOPs & Bacc (CI) \\
    \midrule
    H0-mini~\cite{filiot2025distilling} & 22.3 & 94.0 (92.3--95.4) \\
    UNI2-h~\cite{chen2024towards} & 180.4 & \textbf{95.0} (94.0--96.5) \\
    Vanilla MTD & 21.97 & 93.7 (92.3--95.5) \\
    \textbf{OMNI-B (ours)} & 21.8 & \underline{94.9} (93.3--96.1) \\
    \bottomrule
  \end{tabular}
\end{table}

\section{MoE routing statistics}
\label{sec:moe_routing}

MoE layers are introduced to give the student more flexibility in how it
processes the different token types, in particular the $M{=}10$
teacher-specific tokens. This flexibility is only useful if the router
actually assigns different experts to different tokens, rather than
collapsing onto a single expert pair. We therefore analyze the routing
decisions of our model on $5$K randomly sampled training tiles.

\myparagraph{Protocol} For each image and each of the three MoE layers, we
record the ordered top-2 expert set selected for every token. With $E{=}5$
experts, there are $20$ possible ordered sets. We then compute the largest
number of tokens that share the same set, separately for teacher tokens and
patch tokens. A fully collapsed router would assign the same set to all
tokens. For reference, if the $20$ sets were drawn uniformly at random, the
largest group would typically contain $2$--$3$ of the $10$ teacher tokens,
and about $8\%$ of the $256$ patch tokens.

\begin{figure}[h!]
  \centering
  \includegraphics[width=\linewidth]{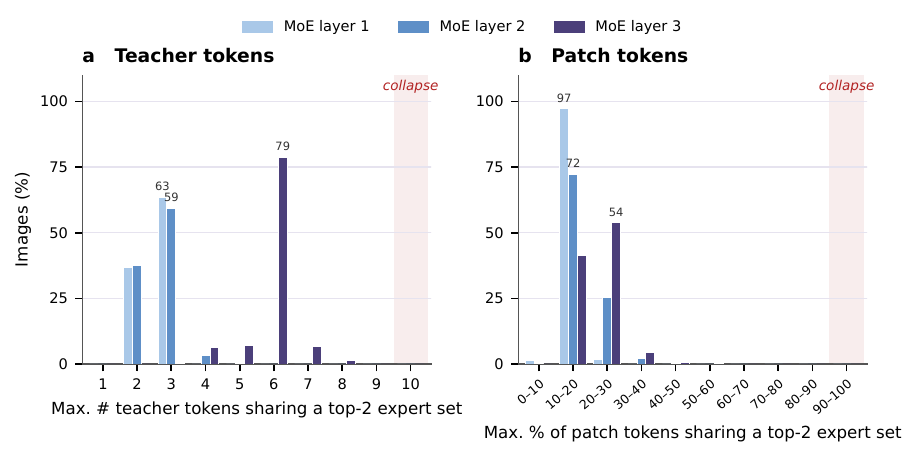}
  \caption{\textbf{MoE routing diversity} of OMNI-B over $5$K random training
  tiles (top-2 routing, $E{=}5$).
  \textbf{(a)} Fraction of images by the maximum number of teacher tokens
  sharing the same ordered top-2 expert set.
  \textbf{(b)} Fraction of images by the maximum percentage of patch tokens
  sharing the same ordered top-2 expert set.
  Shaded bins mark full routing collapse.}
  \label{fig:moe_routing}
\end{figure}

\myparagraph{Teacher tokens} Figure~\ref{fig:moe_routing}a reports the
distribution of this maximum over images. In the first two MoE layers, at
most $2$ or $3$ teacher tokens share the same expert set in more than $96\%$
of images. This is close to the random-assignment reference, so routing is
highly diverse across teachers. The last MoE layer shows more sharing, with a
mode at $6$ teacher tokens ($78.7\%$ of images). Routing nonetheless never
collapses, as no image has all $10$ teacher tokens on the same expert set.
This increased sharing in the deepest layer suggests that groups of related
teachers are handled by common experts close to the output.

\myparagraph{Patch tokens} Figure~\ref{fig:moe_routing}b reports the same
statistic for patch tokens, binned as a percentage of the $256$ patches per
image. In all layers, the most shared expert set covers only $10$--$30\%$ of
patches, far from the collapse regime ($90$--$100\%$). In the first layer,
$97.2\%$ of images fall in the $10$--$20\%$ bin. Sharing increases slightly
with depth, and in the last layer the $20$--$30\%$ bin holds $53.8\%$ of
images.

Overall, both teacher and patch tokens are routed to diverse sets of experts,
and the load-balancing loss $\mathcal{L}_{\mathrm{MoE}}$ prevents the router
from collapsing.

\section{Causes of the scaling failure}
\label{sec:scaling_failure}

Besides the single-CLS bottleneck, optimization difficulty and conflicts
between teachers could also partially explain the scaling failure of vanilla
MTD. We conducted experiments to probe the impact of these two causes, and did
not detect any conflict between the supervision coming from different teachers
or any optimization instability in our context of digital pathology foundation
model teachers. We detail these experiments below.

\myparagraph{Teacher conflicts} To investigate potential teacher conflicts, we
adapt the gradient conflict analysis commonly used in multi-task
learning~\cite{yu2020gradient} to the MTD setting. We quantify conflict by
computing the gradient of each teacher's individual loss with respect to the
student parameters. We then calculate the average pairwise cosine similarity
between each teacher's gradient vector and those of the rest of the pool,
where negative values would indicate actively opposing update directions. As
shown in Table~\ref{tab:grad_cosine}, training the vanilla MTD baseline on our
pathology teachers does not produce major gradient conflicts. Instead, the
update directions remain positively aligned across the ten teachers, with a
mean cosine similarity of $0.162$.

\begin{figure}[h!]
  \centering
  \includegraphics[width=\linewidth]{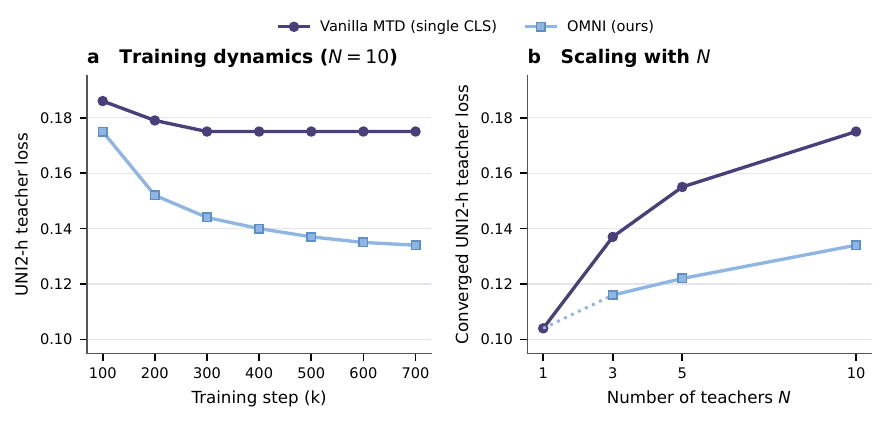}
  \caption{\textbf{UNI2-h teacher reconstruction loss.}
  \textbf{(a)} Loss during training with $N{=}10$ teachers. Both methods
  converge smoothly, with OMNI reaching a lower value.
  \textbf{(b)} Converged loss as a function of the number of teachers $N$. The
  single-CLS bottleneck of vanilla MTD degrades sharply as $N$ grows. With a
  single teacher, both methods reduce to the same model.}
  \label{fig:scaling_failure_loss}
\end{figure}

\begin{table}[ht]
\centering
\caption{Gradient conflict analysis for vanilla MTD (across 500k iterations).}
\begin{tabular}{lccc}
\toprule
Teacher & Mean grad. cosine w/ others & Min & Max \\
\midrule
UNI2-h & 0.229 & 0.049 & 0.418 \\
Virchow2 & 0.137 & 0.035 & 0.206 \\
H-optimus-1 & 0.225 & 0.046 & 0.423 \\
Prov-GigaPath & 0.201 & 0.050 & 0.344 \\
Kaiko-ViT-B/8 & 0.135 & 0.039 & 0.209 \\
Hibou-L & 0.168 & 0.062 & 0.250 \\
H0-mini & 0.233 & 0.090 & 0.423 \\
KEEP & 0.149 & 0.043 & 0.216 \\
DINOv3-ViT-L/16 & 0.050 & 0.035 & 0.090 \\
CONCH1.5 & 0.094 & 0.039 & 0.155 \\
\midrule
\textbf{Overall} & \textbf{0.162} & \textbf{0.035} & \textbf{0.423} \\
\bottomrule
\end{tabular}
\label{tab:grad_cosine}
\end{table}
\myparagraph{Optimization difficulty} To probe for optimization difficulty or
training instability, we tracked the reconstruction loss throughout training.
If optimization difficulties were the primary cause of the scaling failure, we
would expect highly volatile loss curves, spikes, or an inability to converge.
Instead, as shown in Figure~\ref{fig:scaling_failure_loss}a (sampled across
$700$k training steps for the $N{=}10$ case), the loss curves of both vanilla
MTD and our method are smooth and stable. The loss of vanilla MTD converges
faster, but to a higher value.

\myparagraph{Single-CLS bottleneck} Finally, to link the performance
degradation to the CLS token bottleneck, we analyze the converged loss as a
function of the number of teachers $N$. As shown in
Figure~\ref{fig:scaling_failure_loss}b, when scaling from $1$ to $10$ teachers,
the converged loss of vanilla MTD rapidly degrades, increasing from $0.104$ to
$0.175$. A single CLS token lacks the representational capacity to
simultaneously compress the diverse feature distributions of multiple
teachers. By introducing teacher-specific tokens, OMNI reaches a lower
converged loss for every number of teachers considered.

\section{Hyperparameter sensitivity analysis}

Table~\ref{tab:lambda_ctr} presents a sensitivity analysis on the hyperparameter $\lambda_{ctr}$, where values $0$, $0.1$, $0.25$, $0.5$ are compared. As can be seen, removing the loss ($\lambda_{ctr} = 0$) leads to a decrease in performance (as already shown in our ablation study in Table~\ref{tab:ablation}), similarly to fixing a too high weight ($\lambda_{ctr} = 0.5$). Our method is however robust to moderate weight variations with the same performance for $\lambda_{ctr} = 0.1$ and $\lambda_{ctr} = 0.25$.

\begin{table}[ht]
\centering
\caption{Sensitivity to the contrastive loss weight $\lambda_{ctr}$. F1 (\%) averaged across 13 THUNDER tasks.}
\begin{tabular}{cccc}
\toprule
$\lambda_{ctr} = 0$ & $\lambda_{ctr} = 0.1$ & $\lambda_{ctr} = 0.25$ & $\lambda_{ctr} = 0.5$ \\
\midrule
86.6 & \textbf{87.1} & \textbf{87.1} & 86.8 \\
\bottomrule
\end{tabular}
\label{tab:lambda_ctr}
\end{table}

\section{Patch loss details}
\label{sec:patch_loss_details}

\myparagraph{Patch-grid interpolation for heterogeneous teacher tokenizations}
In our multi-teacher setting ($M{=}10$), the number of patch tokens produced by teacher backbones can vary substantially
(e.g., $7{\times}7$, $14{\times}14$, $16{\times}16$, $28{\times}28$), which prevents a direct one-to-one comparison between
student and teacher patch embeddings.
To handle this mismatch, we treat patch tokens as samples on a regular 2D grid and \emph{resize} the student patch-token grid
to the teacher grid using differentiable bilinear interpolation before applying the teacher-specific patch projector.

Let the student produce patch tokens
$\mathbf{Z}^{\mathrm{patch}}_{\theta}(d_j)\in\mathbb{R}^{N_s\times D_s}$ for sample $d_j$,
and teacher $i$ produce patch embeddings
$\mathbf{Y}^{\mathrm{patch}}_{i}(d_j)\in\mathbb{R}^{N_i\times D_i}$.
We assume both $N_s$ and $N_i$ correspond to square grids, i.e.,
$N_s=H_sW_s$ and $N_i=H_iW_i$ with $H_s=W_s=\sqrt{N_s}$ and $H_i=W_i=\sqrt{N_i}$.
We first reshape student tokens into a feature map $\mathbf{F}^{s}_{\theta}(d_j)\in\mathbb{R}^{D_s\times H_s\times W_s}$

Then, for each teacher $i$, we resize this feature map to the teacher grid using bilinear interpolation:
\begin{equation}
\mathbf{F}^{s\rightarrow i}_{\theta}(d_j)
=
\mathrm{Interp}_{H_i\times W_i}\!\left(\mathbf{F}^{s}_{\theta}(d_j)\right)
\in\mathbb{R}^{D_s\times H_i\times W_i}.
\end{equation}
Finally, we flatten back to obtain \emph{teacher-aligned} student tokens $
\bar{\mathbf{Z}}^{\mathrm{patch},i}_{\theta}(d_j) \in\mathbb{R}^{N_i\times D_s}.
$
Dense supervision is provided only on a subset of teacher positions $\mathcal{I}_i$ (precomputed sampled indices),
so the patch-level distillation term becomes
\begin{equation}
\mathcal{L}^{\mathrm{patch}}_{\mathrm{MTD},i}(\theta,\psi_i,\bar{w}_i,d_j)
=
\frac{1}{|\mathcal{I}_i|}\sum_{u\in\mathcal{I}_i}
\Phi\!\left[
p^{\mathrm{patch}}_{\bar{w}_i}\!\left(\bar{\mathbf{z}}^{\mathrm{patch},i}_{\theta,u}(d_j)\right),\
\mathbf{y}^{\mathrm{patch}}_{i,u}(d_j)
\right],
\end{equation}
where $\bar{\mathbf{z}}^{\mathrm{patch},i}_{\theta,u}(d_j)\in\mathbb{R}^{D_s}$
denotes the $u$-th aligned student token on the grid of teacher $i$,
$\mathbf{y}^{\mathrm{patch}}_{i,u}(d_j)\in\mathbb{R}^{D_i}$ the corresponding
teacher patch embedding, and $p^{\mathrm{patch}}_{\bar{w}_i}:\mathbb{R}^{D_s}
\rightarrow\mathbb{R}^{D_i}$ the linear projection head of teacher $i$, applied
independently to each token. $\Phi$ is the cosine and smooth L1 loss defined in
Section~\ref{sec:method}. In practice, several teachers share the same grid
size $N_i$. We therefore compute
$\bar{\mathbf{Z}}^{\mathrm{patch},i}_{\theta}(d_j)$ once per distinct grid size
and reuse it across teachers for efficiency. The final patch loss of teacher
$i$ is obtained by averaging this term over the samples of the batch.

\section{Implementation details}
\label{sec:implem_details}

We provide here additional implementation details for reproducibility.
All models are trained from random initialization on $10$M tiles extracted
from $10$K whole-slide images sourced from TCGA, spanning diverse tissue
types and organ systems. Teacher embeddings are pre-computed and stored
offline; teachers are kept fully frozen throughout student training.
Table~\ref{tab:hyperparameters} summarizes the main hyperparameters for
all OMNI variants.
\begin{table}[h]
\centering
\caption{Training hyperparameters for OMNI models.}
\label{tab:hyperparameters}
\begin{tabular}{lcccc}
\toprule
\textbf{Hyperparameter} & \textbf{OMNI-T} & \textbf{OMNI-S} & \textbf{OMNI-B} & \textbf{OMNI-L} \\
\midrule
\multicolumn{5}{l}{\textit{Optimization}} \\
\quad Optimizer & \multicolumn{4}{c}{Adam} \\
\quad Learning rate & \multicolumn{4}{c}{$1 \times 10^{-4}$} \\
\quad Training iterations & \multicolumn{4}{c}{700k} \\
\quad Batch size & \multicolumn{4}{c}{128} \\
\midrule
\multicolumn{5}{l}{\textit{Architecture}} \\
\quad Embedding dimension & 192 & 384 & 768 & 1024 \\
\quad Depth & 12 & 12 & 12 & 24 \\
\quad Attention heads & 3 & 6 & 12 & 16 \\
\quad Parameters & 7.3M & 28.7M & 114M & 330M \\
\quad MoE tail blocks & 3 & 3 & 3 & 5 \\
\quad Number of experts $E$ & \multicolumn{4}{c}{5} \\
\quad Top-$k$ routing & \multicolumn{4}{c}{$k = 2$} \\
\quad MoE auxiliary loss coefficient $\alpha_\mathrm{MoE}$ & \multicolumn{4}{c}{0.05} \\
\quad Image size & \multicolumn{4}{c}{$224 \times 224$} \\
\quad Patch size & \multicolumn{4}{c}{$14 \times 14$} \\
\midrule
\multicolumn{5}{l}{\textit{Loss}} \\
\quad CLS loss & \multicolumn{4}{c}{Cosine + SmoothL1 ($\beta{=}1.0$)} \\
\quad SmoothL1 weight $\lambda_\mathrm{reg}$ & \multicolumn{4}{c}{1.0} \\
\quad Contrastive loss & \multicolumn{4}{c}{InfoNCE} \\
\quad Contrastive weight $\lambda_\mathrm{ctr}$ & \multicolumn{4}{c}{0.1} \\
\quad Contrastive temperature $\tau$ & \multicolumn{4}{c}{0.1} \\
\quad Patch loss & \multicolumn{4}{c}{\checkmark} \\
\midrule
\multicolumn{5}{l}{\textit{Data}} \\
\quad Pre-training dataset & \multicolumn{4}{c}{TCGA} \\
\quad Number of tiles & \multicolumn{4}{c}{10M} \\
\quad Number of WSIs & \multicolumn{4}{c}{10k} \\
\quad Number of teachers $M$ & \multicolumn{4}{c}{10} \\
\bottomrule
\end{tabular}
\end{table}

\section{Downstream evaluation details}
\label{sec:downstream_details}

We provide here additional details on the downstream evaluation benchmarks
used in this work.

\myparagraph{Slide-level evaluation protocol} For slide-level evaluation, we
follow the PathoBench protocol~\cite{zhang2025accelerating} on CPTAC cohorts,
and report balanced accuracy (bacc) over a 50-fold train/test split. The full
list of $25$ tasks across $7$ cancer types is provided in
Table~\ref{tab:slide_tasks}.

\myparagraph{Mean pooling as the default aggregation choice} We use
mean-pooling over teacher-specific tokens as our default aggregation method
for two main reasons. First, it isolates the gains brought by our pre-training
contributions from those of any additional downstream adaptation. This allows
us to clearly measure the benefits of our teacher-specific formulation and its
accompanying architectural choices, and to show that no extra downstream
computation is needed to improve upon vanilla MTD. Second, mean-pooled
embeddings can be computed once, offline, before training a task-specific
probe. This is particularly interesting for MIL models: learning per-teacher
weights would instead require storing all teacher-specific embeddings for
every patch, multiplying storage requirements by $M{=}10$. It would also
require loading these embeddings into GPU memory during training, which slows
down training and increases memory constraints. Mean-pooling avoids both
issues, which is why we adopt it as our default setting.

Our method combined with simple and efficient mean-pooling is thus sufficient
to better scale to more teachers. Performance can nonetheless be further
improved with a learned aggregation: as shown in Table~\ref{tab:bracs_slide},
a weighted pooling of teacher-specific tokens brings additional gains on the
BRACS dataset.

\section{Compute budget summary}
\label{sec:sup_compute_budget}

Pre-training consists of two stages. First, we pre-computed and stored offline the embeddings of all $10$ teacher models over the $10$M-tile TCGA dataset, requiring approximately $400$ A100 GPU hours in total. Each subsequent student pre-training run then took approximately $40$ H100 GPU hours. For downstream evaluation, benchmarking on THUNDER and PathoBench required approximately $10{,}000$ V100 GPU hours in total, the majority coming from tile feature extraction across the large number of whole-slide images for all foundation models.

\begin{table}[h]
\centering
\caption{Slide-level tasks from the PathoBench benchmark~\cite{zhang2025accelerating} on CPTAC cohorts. All tasks are evaluated using balanced accuracy (BACC) with a 50-fold train/test split.}
\label{tab:slide_tasks}
\resizebox{\linewidth}{!}{%
\begin{tabular}{llll}
\toprule
\textbf{Cohort} & \textbf{Cancer Type} & \textbf{Task} & \textbf{Task Type} \\
\midrule
CPTAC-BRCA & Breast carcinoma          & Immune class         & Subtyping \\
            &                           & PIK3CA mutation      & Mutation prediction \\
            &                           & TP53 mutation        & Mutation prediction \\
\midrule
CPTAC-CCRCC & Clear cell renal cell carcinoma & BAP1 mutation  & Mutation prediction \\
             &                                 & VHL mutation   & Mutation prediction \\
\midrule
CPTAC-COAD  & Colon adenocarcinoma      & ACVR2A mutation      & Mutation prediction \\
             &                           & ARID1A mutation      & Mutation prediction \\
             &                           & Immune class         & Subtyping \\
             &                           & MSI-H                & Biomarker prediction \\
             &                           & PIK3CA mutation      & Mutation prediction \\
             &                           & SETD1B mutation      & Mutation prediction \\
             &                           & TP53 mutation        & Mutation prediction \\
\midrule
CPTAC-GBM   & Glioblastoma multiforme   & EGFR mutation        & Mutation prediction \\
             &                           & Immune class         & Subtyping \\
             &                           & TP53 mutation        & Mutation prediction \\
\midrule
CPTAC-LSCC  & Lung squamous cell carcinoma & ARID1A mutation   & Mutation prediction \\
             &                              & Histologic grade  & Grading \\
             &                              & Immune class      & Subtyping \\
             &                              & KEAP1 mutation    & Mutation prediction \\
\midrule
CPTAC-LUAD  & Lung adenocarcinoma       & EGFR mutation        & Mutation prediction \\
             &                           & KRAS mutation        & Mutation prediction \\
             &                           & STK11 mutation       & Mutation prediction \\
             &                           & TP53 mutation        & Mutation prediction \\
\midrule
CPTAC-UCEC  & Uterine corpus endometrial carcinoma & CTNNB1 mutation & Mutation prediction \\
             &                                      & PTEN mutation   & Mutation prediction \\
\midrule
\multicolumn{2}{l}{\textbf{Total}} & \textbf{25 tasks} & \textbf{7 cohorts} \\
\bottomrule
\end{tabular}}
\end{table}

\clearpage
\newpage

\end{document}